\documentclass[acmtog]{acmart}

\usepackage{graphicx}
\usepackage{amsmath}
\usepackage{caption}
\usepackage{multirow}
\usepackage{url}
\usepackage{color}
\usepackage{comment}
\usepackage{enumitem}
\usepackage{xspace}

\let\emptyset\varnothing

\usepackage{pifont}

\def\eg{\emph{e.g.,}\xspace}
\def\ie{\emph{i.e.,}\xspace}

\def\etal{\emph{et al.}\xspace}

\AtBeginDocument{%
  }

\copyrightyear{2026}
\acmYear{2026}
\setcopyright{cc}
\setcctype{by-nc-nd}
\acmConference[SA Conference Papers '26]{SIGGRAPH Asia 2026 Conference Papers}{December 01--04, 2026}{Kuala Lumpur, Malaysia}
\acmBooktitle{SIGGRAPH Asia 2026 Conference Papers (SA Conference Papers '26), December 01--04, 2026, Kuala Lumpur, Malaysia}
\acmDOI{10.1145/3829340.3842350}
\acmISBN{979-8-4007-2842-6/2026/12}

\acmSubmissionID{2384}

\begin{document}

\title{DCGrasp: Distance-aware Controllable Grasp Generation} 


\author{Hiroyasu Akada}
\affiliation{%
  \institution{Max Planck Institute for Informatics}
  \country{Germany,}
    \institution{and Google}
  \country{Switzerland}
  }
\email{hakada@mpi-inf.mpg.de}

\author{Jesús Pérez}
\affiliation{%
  \institution{Google}
  \country{Switzerland}}
\email{jesusprod@google.com}

\author{Emre Aksan}
\affiliation{%
  \institution{Google}
  \country{Switzerland}}
\email{emre.aksn@gmail.com}

\author{Vasileios Choutas}
\affiliation{%
  \institution{Google}
  \country{Switzerland}}
\email{vchoutas@google.com}

\author{Cristian Romero}
\affiliation{%
  \institution{Independent Researcher}
  \country{Spain}}
\email{crisrom002@gmail.com}

\author{Alberto Garcia-Garcia}
\affiliation{%
  \institution{Google}
  \country{Switzerland}}
\email{bergo@google.com}

\author{Vladislav Golyanik}
\affiliation{%
  \institution{Max Planck Institute for Informatics}
  \country{Germany}}
\email{golyanik@mpi-inf.mpg.de}

\author{Christian Theobalt}
\affiliation{%
  \institution{Max Planck Institute for Informatics}
  \country{Germany}}
\email{theobalt@mpi-inf.mpg.de}

\author{Thabo Beeler}
\affiliation{%
  \institution{Google}
  \country{Switzerland}}
\email{thabo.beeler@gmail.com}





\renewcommand{\shortauthors}{Akada et al.}

%
%
\begin{abstract}
    Generating 3D hand-object interactions is essential for applications in robotics, AR/VR, and synthetic data generation, where flexible controllability and strong generalization to diverse object geometries are required.
    However, existing methods rarely satisfy these requirements, limiting their practical applicability.
    We present DCGrasp, a multi-modal diffusion framework for distance-aware controllable grasp generation.
    DCGrasp is built on a novel grasp energy term that computes \textit{Distance Profile}, a signed distance from each hand vertex to the nearest object point, coupled with distance-aware weighting, effectively capturing the semantically similar hand–object interaction in near-contact regions while remaining invariant to object and hand identity.   %
    Given various controllable signals, DCGrasp first generates a Distance Profile based on a Diffusion Transformer, together with a corresponding candidate hand pose.
    We then refine the candidate pose through optimization, enforcing consistency between the optimized hand pose and the generated Distance Profile in near-contact regions.
    Our experiments show that DCGrasp produces high-quality, geometrically plausible grasps with flexible user control, generalizing to diverse object and hand shapes and scales.
    Our work establishes a robust and versatile pipeline for the synthesis of controllable 3D hand–object interactions.    
    The supplementary video with dynamic visualization is available on our project page\footnote{\url{https://4dqv.mpi-inf.mpg.de/DCGrasp/}}.
\end{abstract}





\begin{teaserfigure}
  \centering
  \includegraphics[width=1.0\linewidth]{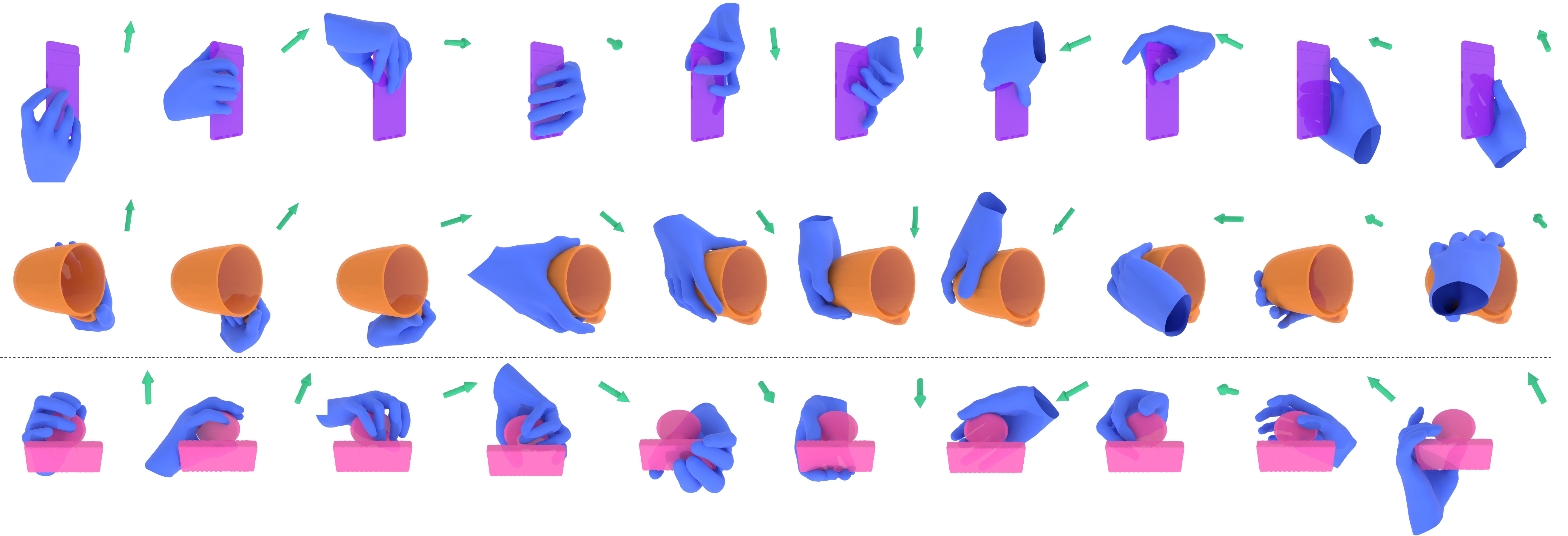}
  \vspace{-6mm}
  \caption{Qualitative results for various hand direction signals. Our framework effectively produces physically plausible results complying with the conditioning signal with out-of-domain objects. 
  }
  \label{fig:qualitative_results_hdv}
\end{teaserfigure}


\maketitle



%
%
\section{Introduction}
\label{sec:intro}
Synthesizing realistic 3D hand–object interactions (HOI) is fundamental to applications such as robotics, AR/VR, and synthetic data generation. 
These applications require not only geometrically plausible grasps but also flexible controllability and strong generalization to a variety of object geometries and scales.
However, existing methods do not fully satisfy these requirements. 
Controllability is largely underexplored in the HOI literature, with limited support for explicit conditioning signals.
Moreover, learning-based approaches often struggle to generalize to different hand identities and novel objects with diverse shapes and scales outside of the training datasets.
These shortcomings largely limit their practical applicability.

\begin{figure*}[tb]
    \centering
      \includegraphics[width=1.0\linewidth]{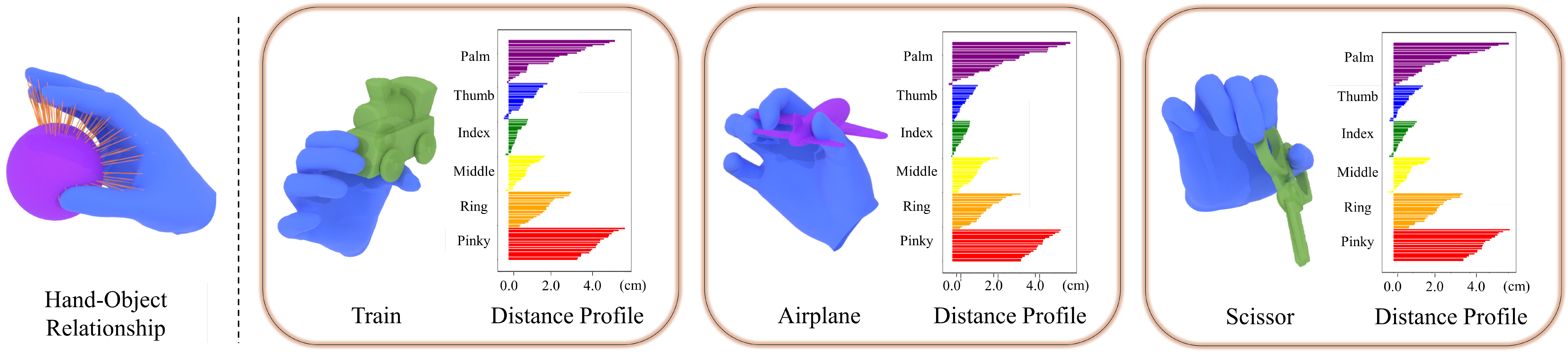}
      \vspace{-4mm}
      \caption{\textit{Left}: Distance Profiles are computed as the signed distance from a selection of hand surface vertices (the inner palm
and finger surfaces) \textit{Right}: 
      A set of interactions yielding similar Distance Profiles, which preserves overall structural semantics across diverse hand-object grasp pairs.
      }
      \label{fig:teaser}
    \vspace{-1mm}
\end{figure*}

In this paper, we propose \textit{DCGrasp}, 
a multi-modal diffusion framework for \textbf{D}istance-aware and \textbf{C}ontrollable \textbf{Grasp} synthesis.
DCGrasp supports multiple conditioning signals, including object geometry, hand shape, orientation, and root distance.
Achieving robust generalization across such a diverse input domain constitutes a great challenge with publicly available training data. Moreover, capturing hand-object interactions is time-consuming, involving non-trivial mesh reconstruction and object tracking. Therefore, we aim to improve cross-object generalization by exploring alternative grasp representations.

In particular, we observe that Euclidean distances remain similar across near-contact geometries for the groups of semantically close grasping interactions, as illustrated in  Fig.~\ref{fig:teaser}.
Motivated by this insight, we build our framework upon a novel grasp energy term.
This term comprises two concepts: \textit{Distance Profile} and distance-aware weighting.
Distance Profile encodes the signed distance from each hand vertex to its nearest object point.
Despite its simplicity, we find that Distance Profile captures hand-object grasp patterns in a manner that is invariant to geometry while preserving the overall structural semantics: \textit{poke}, \textit{touch}, \textit{grab}, \textit{pinch}, etc.
We then introduce a distance-aware weighting, which prioritizes interaction-critical geometries while reducing the influence of distant, non-interacting regions.
Based on this grasp energy term, our framework exhibits strong 
cross-object generalization, enabling the synthesis of high-quality grasps for unseen or arbitrarily scaled hands and objects.

DCGrasp consists of two stages: (1) Diffusion–based Distance Profile generation and (2) grasp energy-based optimization. 
Firstly, given various controllable signals, our Diffusion Transformer model generates a Distance Profile together with a corresponding candidate hand pose.
Secondly, we refine the candidate pose via grasp energy–based optimization, enforcing consistency between the synthesized pose and the generated Distance Profile in near-contact regions.
This two-stage design decouples interaction modeling from object-specific geometry, leading to improved controllability and robustness to a wide variety of objects.
Unlike existing methods~\cite{contactopt, jiang2021handobjectcc} that typically require two separate models to generate initial poses and contact maps for subsequent optimization (\ie three-stage pipelines), 
DCGrasp employs a unified model that jointly produces initial poses and target profiles, simplifying the system training.

Our experiments show that DCGrasp produces high-quality, semantically consistent grasps that generalize effectively to unseen object geometries and scales.
Our method establishes a robust pipeline for physics-plausible and user-controllable hand–object interaction synthesis (Fig.~\ref{fig:qualitative_results_hdv}).
In summary, our contributions are threefold:

\begin{itemize} 
    \setlength{\itemsep}{1pt} 
    \item \textbf{A distance-aware grasp energy term}: Our energy term combines Distance Profile with a distance weighting scheme, producing families of semantically similar grasps in near-contact regions while invariant to geometries.
    \item \textbf{A controllable grasp generation system}: We propose \textit{DCGrasp}, a two-stage pipeline that combines Diffusion-based Distance Profile generation with Distance Profile-based optimization, allowing explicit conditioning on object geometry, hand shape, orientation, and root distance. 
    \item \textbf{Generalization to new hands, object shapes, and scales}:  DCGrasp decouples interaction modeling from object-specific geometry, producing high-quality, geometrically plausible grasps that effectively generalize to new object shapes and scales.
\end{itemize}

%
%
\section{Related Work}
\label{sec:related_work}

\subsection{Hand-Object Interaction Representations.}
%
%
Recent years have seen substantial progress in hand-object interaction, driven by the development of various hand models. 
Early works explicitly model 3D hand shape with meshes~\cite{hand_primitives, hand_meshes} and  
parametric models~\cite{mano,nimble}.
Other works use implicit functions to represent hand and object relationships, such as 
3D Gaussians~\cite{hand_gaussians} or 3D distance fields~\cite{grasping_field, lisa} that requires computing Marching Cubes~\cite{marchingcubes} using a large number of 3D point queries.
For example, GRIP~\cite{taheri_3dv2024_grip} uses sensor-based distance features as network inputs for grasp generation and refinement, while She~\etal~\cite{10.1145/3528223.3530091} adopt Interaction Bisector Surface~\cite{10.1145/2574860}, a 3D geometric interaction representation, as the RL state observation for gripper-object interaction. In both methods, the interaction representation is used as an input rather than being learned or predicted.
D(R,O) Grasp~\cite{wei2024dro} is closer to our work, predicting a dense point-to-point pairwise distance matrix, which is first decoded into target poses, and optimization reconstructs a kinematically feasible robot configuration.
In contrast, DCGrasp learns a 1D signed per-hand-vertex Distance Profile that is consistent with the controllable signals. The predicted Distance Profile is directly used through our grasp-energy term to refine a generated grasp. Thus, Distance Profile serves as a guidance signal specifying the desired hand-object interaction, rather than being decoded into hand geometry.
This formulation captures the main structural semantics of near-contact interactions while remaining agnostic to hand-object relative poses. 
%

\subsection{Hand-Object Interaction Datasets.}
%
%
Following the diverse hand model formats, many datasets have been proposed using single~\cite {contactpose, dexYCB, fhpa, honnotate, freihand, hoi4d} or two-hand~\cite {h2o, h2o-3d} images.
Some work captures hand meshes~\cite{smplx} that grasp rigid~\cite{grab, liu2024taco} or articulated objects~\cite{arctic, Liu_2022_CVPR}. 
Recent work generates synthetic HOI datasets using a diffusion model~\cite{hoidiffusion} or an optimization approach~\cite{dexgraspnet}.
Although these existing datasets cover diverse object categories, their hand shape variations are limited (\eg 10 subjects~\cite{grab, arctic}).
Therefore, we collected a large-scale HOI dataset with 100 subjects for a thorough evaluation of our framework that supports hand shape conditioning.
However, object diversity is still limited (5 total objects), and thus our distance-aware grasp energy term remains critical to achieve cross-object generalization.

\begin{figure*}[t]
 \centering
    \includegraphics[width=1.0\linewidth]{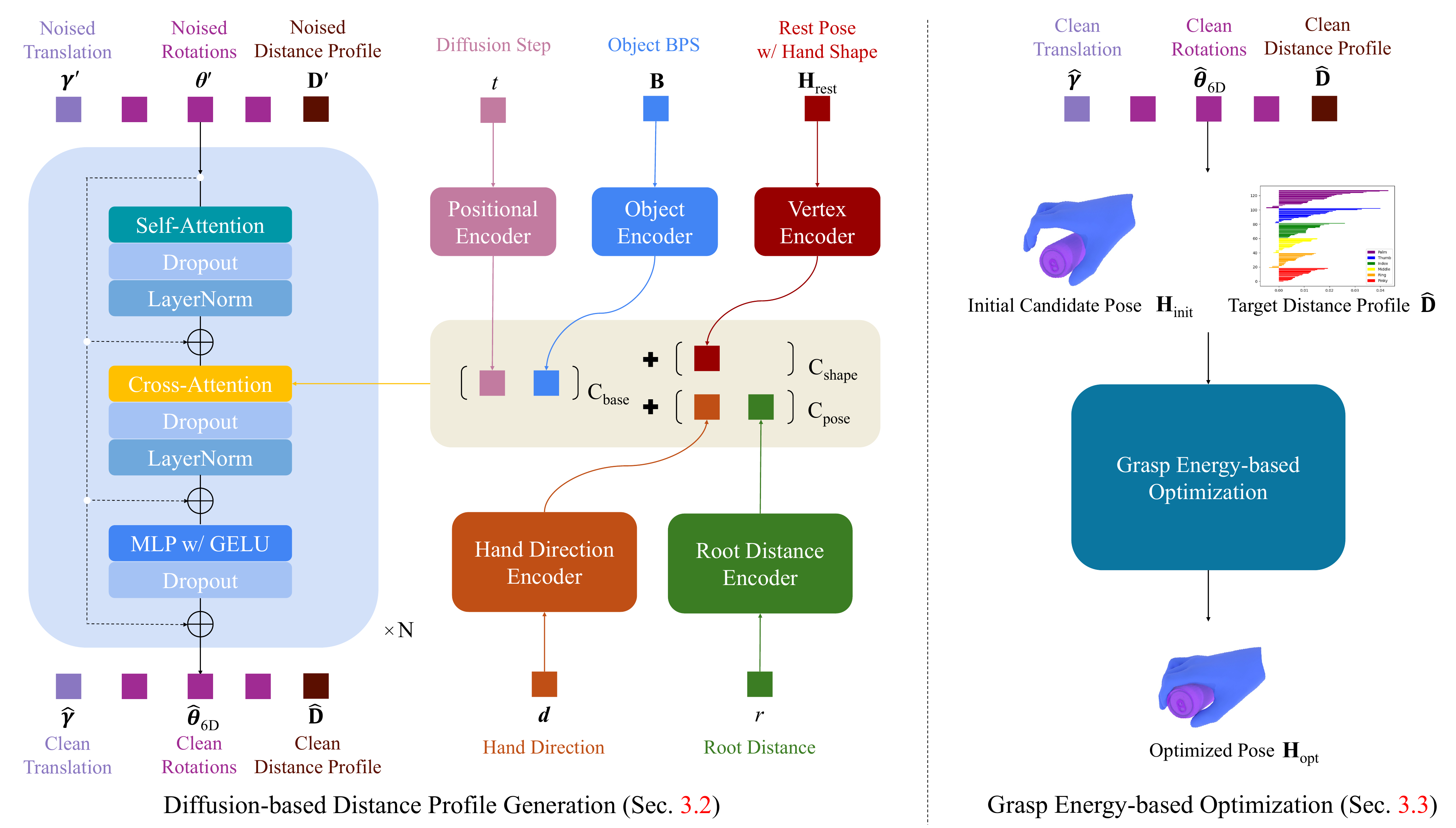}
 \caption{
 \textbf{Method overview.} Given object BPS and a series of control signals, our diffusion model first generates a Distance Profile along with the parameters of an initial candidate pose (Sec.~\ref{sec:method_generation}).
 We then refine the parameters of the candidate pose via grasp energy–based optimization, enforcing consistency between the synthesized hand pose and the generated profile in near-contact regions (Sec.~\ref{sec:method_optimization}).
 }
 \label{fig:overview_method} 
\end{figure*}

\subsection{Grasp Generation.}
%
%
Existing works can be broadly categorized into two groups: hand-only grasp generation and whole-body generation with hands.

The first group includes grasp estimation from images~\cite{hasson2019obman, ganhand, fan2024hold, swamy2023showme, ye2023ghop}, grasp motion synthesis~\cite{dgrasp, manipnet, geneoh, taheri_3dv2024_grip, graspxl, toch, hao2024hand, zhang2025bimart, zhang2025manidext, cha2024text2hoi, MACS2024, Zheng_2023_CVPR}, two-hand interaction~\cite{zhang2025bimart, zhang2025manidext, cha2024text2hoi, lee2024interhandgen, MACS2024, zhang2024artigrasp, Muchen_LatentHOI}, and robotic hands grasping~\cite{huang2023dynamic, lee2024dextouch, turpin2023fastgraspd, wan2023unidexgrasp++, wang2024cyberdemo, wang2022dexgraspnet, touch-dexterity, murali2025graspgen}. 
In particular, Cao~\etal~\cite{cao2025joint} tackles multi-modal grasp generation. However, its framework requires additional large-scale object data for training and does not support the controllability discussed in Sec.~\ref{sec:intro}.
Contact-based approaches~\cite{contactgrasp, contactopt, jiang2021handobjectcc, turpin2022graspddc, contactgen, grab, zhang2025bimart, cha2024text2hoi} also generate promising grasps by explicitly modeling physical interaction, \eg leveraging a real contact dataset~\cite{contactdb} or predicted contact maps, as in ContactOpt~\cite{contactopt} and GraspTTA~\cite{jiang2021handobjectcc}. 
These approaches generally rely on a three-stage pipeline: an initial hand pose is first generated, followed by contact prediction and subsequent optimization.
In contrast, our grasp energy-based approach allows for a unified formulation that jointly produces both initial poses and target interaction profiles.
%

The second group focuses on the whole-body generation with hands~\cite{flex, paschalidis2025cwgrasp, circle, diomataris2024wandr, goal, saga, imos, omomo}.
Many of these works generate poses by optimizing full-body configurations to align with a guiding grasp hand or wrist pose.
Such is the case of FLEX~\cite{flex} and CWGrasp~\cite{paschalidis2025cwgrasp}, for static poses, and CIRCLE~\cite{circle} and WANDR~\cite{diomataris2024wandr} for body motions.
Other works~\cite{goal, saga} directly generate full-body motions that approach an object and converge to a static end grasp pose through optimization, also in a stochastic way~\cite{saga}. 

Among the above works, only a limited number of grasp generation methods explicitly address grasp controllability, primarily focusing on hand direction control via reinforcement learning~\cite{graspxl} or conditioning on hand direction vectors~\cite{paschalidis2025cwgrasp}.
In contrast, our method extends beyond directional control and provides a broader range of user-controllable interaction attributes, while achieving robust generalization to out-of-domain object shapes and scales.

%
%
\section{Method}

%
%
%
%
%
%
\subsection{Hand and Object Representation}
\label{subsec:representation}

\subsubsection{Hand Representation.}
%
%
We follow a parametric model~\cite{mano} $\mathcal{H}(\hat{\boldsymbol{\gamma}}, \hat{\boldsymbol{\theta}}, \boldsymbol{\beta})$ where $\mathcal{H}(\cdot)$ denotes the forward kinematics and mesh deformation function,
parameterized by translation ${\boldsymbol \gamma} \in \mathbb{R}^{3}$, pose ${\boldsymbol \theta} \in \mathbb{R}^{J \times 3}$ encoded as axis-angle rotations, and shape ${\boldsymbol \beta} \in \mathbb{R}^{I}$, where $J$ and $I$ denote the number of hand joints and the dimensionality of the shape space, respectively.
These parameters define a 3D hand mesh $\mathbf{H} \in \mathbb{R}^{S \times 3}$ with normals $\mathbf{N}_{\text{hand}} \in \mathbb{R}^{S \times 3}$ where $S$ is the number of hand vertices.
For training, we use global translation and rotations in an object-centric coordinate, with the rotations converted from axis-angle to 6D representation ${\boldsymbol \theta}_{\text{6D}} \in \mathbb{R}^{J \times 6}$.
As one of the conditioning signals, we introduce a hand direction vector ${\boldsymbol d} \in \mathbb{R}^3$.
To obtain the hand direction vector, we first pre-select two fixed neighboring vertices located on the outer palm of the hand mesh. 
Connecting these two vertices forms a direction vector, which is oriented to align with the anatomical direction from the wrist toward the base of the middle finger.
Note that our hand representation is quasi-rigid around the outer palm area, and thus these vertices stay consistent across poses.

\subsubsection{Object Representation.}
%
%
We follow prior works~\cite{grab, paschalidis2025cwgrasp, zhang2025bimart, omomo} to sample Basis Point Sets (BPS)~\cite{bps} $\mathbf{B} \in \mathbb{R}^{4096}$ uniformly distributed within a sphere of 150 mm radius, without any scale normalization to the objects.
For optimization, we use object point clouds $\mathbf{O} \in \mathbb{R}^{P \times 3}$ with their corresponding surface normals $\mathbf{N}_{\text{obj}} \in \mathbb{R}^{P \times 3}$.

\subsubsection{Interaction Representation.}
%
%
To model the spatial relationship of the hand-object interaction, we adopt a simple representation, \ie \textit{Distance Profile}.
Specifically, we construct a Distance Profile $\mathbf{D} \in \mathbb{R}^{k}$ by pre-sampling $k$ vertices on the hand mesh (covering the inner palm and finger surfaces, where hand-object contacts frequently happen) with the same indices for all hand shapes, and computing the signed distance between each hand vertex and its nearest object point.
The sign is determined by the dot product between the hand-to-object vector and the surface normal at the hand vertex: the distance is positive if the dot product is positive and negative otherwise.
We intentionally adopt this one-sided signed distance (hand-to-object) rather than a symmetric representation to make the representation hand-centric and less dependent on the object geometry, thereby improving generalization to unseen object geometries.
%
%
%
In addition, we incorporate the distance between the root joint and its nearest object point $r \in \mathbb{R}$ as an additional controllable signal.

%
%
%
%
%
%
\subsection{Distance Profile Generation Model}
\label{sec:method_generation}
\subsubsection{Conditional Diffusion Transformer.}
%
%
Given the object feature $\mathbf{B}$ and a set of controllable conditioning signals, our model is designed to generate a Distance Profile along with the parameters of an initial candidate grasp ($\mathbf{D}$, $ \boldsymbol{\gamma} $, $\boldsymbol{\theta}_{\text{6D}}$),
as illustrated in Fig.~\ref{fig:overview_method}.
Here, we adopt a transformer decoder architecture as our diffusion denoiser $M$.

The denoiser first employs task-specific MLP encoders to encode noisy representations ($\mathbf{D}'$, ${\boldsymbol \gamma}'$, $\boldsymbol \theta'_{\text{6D}}$) into latent features $\mathbf{Z}_{\mathbf{D'}} \in \mathbb{R}^{Q}$, $\mathbf{Z}_{\boldsymbol{\gamma'}}  \in \mathbb{R}^{Q}$, and $\mathbf{Z}_{\boldsymbol \theta'_{\text{6D}}}  \in \mathbb{R}^{J \times Q}$, where $Q$ denotes the latent dimension.
Second, these features are concatenated into $Z \in \mathbb{R}^{(J+2) \times Q}$ with $J + 2$ tokens, and interact with each other by a self-attention layer.
Next, the output features from the self-attention layer are processed with the following conditioning modalities: The object BPS $\mathbf{B}$, shape parameters ${\boldsymbol \beta}$, hand direction vector ${\boldsymbol d}$, root distance $r$, and the denoising timestep $t$.
Specifically, these conditioning data are encoded by their corresponding MLP encoders into the same latent space.
Here, the shape parameters ${\boldsymbol \beta}$ are converted into a hand rest pose $\mathbf{H}_{\text{rest}} \in \mathbb{R}^{S \times 3}$ as an input to the encoder.
Then, the self-attention features and conditioning embeddings are injected through a cross-attention layer, enabling the model to learn the corresponding conditional distributions.
Finally, we use task-specific MLP heads to construct model outputs: the Distance Profile $\hat{\mathbf{D}}$, translation ${\hat{\boldsymbol \gamma}}$, and rotation $\hat{\boldsymbol \theta}_{\text{6D}}$.
See the supplement for additional architectural details.

\subsubsection{Diffusion Formulation.}
%
%
We adopt the DDPM formulation and train the Conditional Diffusion Transformer with $\mathbf{X}_0$-prediction parameterization.
The forward diffusion process is defined as:
\begin{equation}
\mathbf{X}_t
=
\sqrt{\bar{\alpha}_t}\mathbf{X}_0
+
\sqrt{1-\bar{\alpha}_t}\boldsymbol{\epsilon},
\quad
\boldsymbol{\epsilon} \sim \mathcal{N}(\mathbf{0}, \mathbf{I}),
\end{equation}
where $\bar{\alpha}_t = \prod_{s=1}^{t} \alpha_s$.
Given the noisy sample $\mathbf{X}_t$ at timestep $t$ and conditions $\mathbf{C}$, the denoiser $M$ predicts the clean representation:
\begin{equation}
\hat{\mathbf{X}}_0
=
M(\mathbf{X}_t, t, \mathbf{C}).
\end{equation}
By decomposing the prediction $\hat{\mathbf{X}}_0 = [ \, \hat{\mathbf{D}} \,|\, \hat{\boldsymbol{\gamma}} \,|\, \hat{\boldsymbol{\theta}}_{\text{6D}} \, ]$, we define the training objective:
\begin{equation}
\mathcal{L}_{\text{total}}
=
\lambda_{\text{dp}}
\left\|
\hat{\mathbf{D}}
-
\mathbf{D}
\right\|_2^2
+
\lambda_{\text{trans}}
\left\|
\hat{\boldsymbol{\gamma}}
-
\boldsymbol{\gamma}
\right\|_2^2
+
\lambda_{\text{rot}}
\left\|
\hat{\boldsymbol{\theta}}_{\text{6D}}
-
\boldsymbol{\theta}_{\text{6D}}
\right\|_2^2,
\end{equation}
where $\lambda_{\text{dp}}$, $\lambda_{\text{trans}}$, and $\lambda_{\text{rot}}$ denote weighting scales. 

\subsubsection{Diffusion Guidance.}
%
%
Classifier-free guidance (CFG)~\cite{ho2022classifier} is a common technique to improve the quality of generated data and to make sampled data better correspond with their conditions.
CFG is implemented by jointly training the model in both conditional and unconditional settings. During inference, the outputs from these two branches are combined at each denoising timestep:
\begin{equation} 
\label{eq:cfg}
\hat{\mathbf{X}}_{0} = M(\mathbf{X}_t, t, \emptyset) + \lambda ( M(\mathbf{X}_t, t, \mathbf{C}) - M(\mathbf{X}_t, t, \emptyset) ),
\end{equation}
where $\lambda$ denotes a guidance scale and $\emptyset$ denotes a null condition.
At inference time, setting $\lambda \geq 1$ steers the sampling trajectory toward the conditional prediction while moving it away from the unconditional one, thereby enhancing condition fidelity. 

In our framework, we integrate a multi-modal classifier-free guidance strategy~\cite{brooks2023instructpix2pix} to accommodate multiple conditioning signals.
Specifically, we partition the conditions into three groups:
\begin{itemize}
    \setlength{\itemsep}{1pt}
    \item $\mathbf{C}_{\text{base}}$, which includes the object BPS $\mathbf{B}$.
    \item $\mathbf{C}_{\text{shape}}$, which includes the shape parameters ${\boldsymbol \beta}$.
    \item $\mathbf{C}_{\text{pose}}$, which includes the input hand direction vector ${\boldsymbol d}$ and the root distance $r$.
\end{itemize}
Where $\mathbf{C}_{\text{base}}$ is always present, and the other two are randomly masked out during training.
Thus, the full conditional prediction is denoted as
$\hat{\mathbf{X}}_0^{\text{full}} 
= M(\mathbf{X}_t, t, \mathbf{C}_{\text{base}}, \mathbf{C}_{\text{shape}}, \mathbf{C}_{\text{pose}})
$.
Based on the defined condition groups, we construct the model output as follows:
\begin{equation} 
\begin{split}
\hat{\mathbf{X}}_{0} & = \ M(\mathbf{X}_t, t, \mathbf{C}_{\text{base}}, \emptyset, \emptyset) \\ 
                     & + \lambda_{\text{shape}} ( M(\mathbf{X}_t, t, \mathbf{C}_{\text{base}}, \mathbf{C}_{\text{shape}}, \emptyset) - M(\mathbf{X}_t, t, \mathbf{C}_{\text{base}}, \emptyset, \emptyset)) \\
                     & + \lambda_{\text{pose}} ( M(\mathbf{X}_t, t, \mathbf{C}_{\text{base}}, \mathbf{C}_{\text{shape}}, \mathbf{C}_{\text{pose}}) \\ 
                     & - M(\mathbf{X}_t, t, \mathbf{C}_{\text{base}}, \mathbf{C}_{\text{shape}}, \emptyset)),
\end{split}
\end{equation}
where $\lambda_{\text{shape}}$ and $\lambda_{\text{pose}}$ are guidance scales for the shape and pose-related conditions, respectively.
This formulation enables flexible control over hand shape and interaction-related signals during sampling. 
We also integrate additional guidance on the Distance Profile to regularize the denoising process, and adopt DDIM~\cite{song2020denoising} for faster sampling.
See our supplement for more details.
%

%
%
%
%

\begin{figure}[tb]
    \centering
    \includegraphics[width=1.0\linewidth]{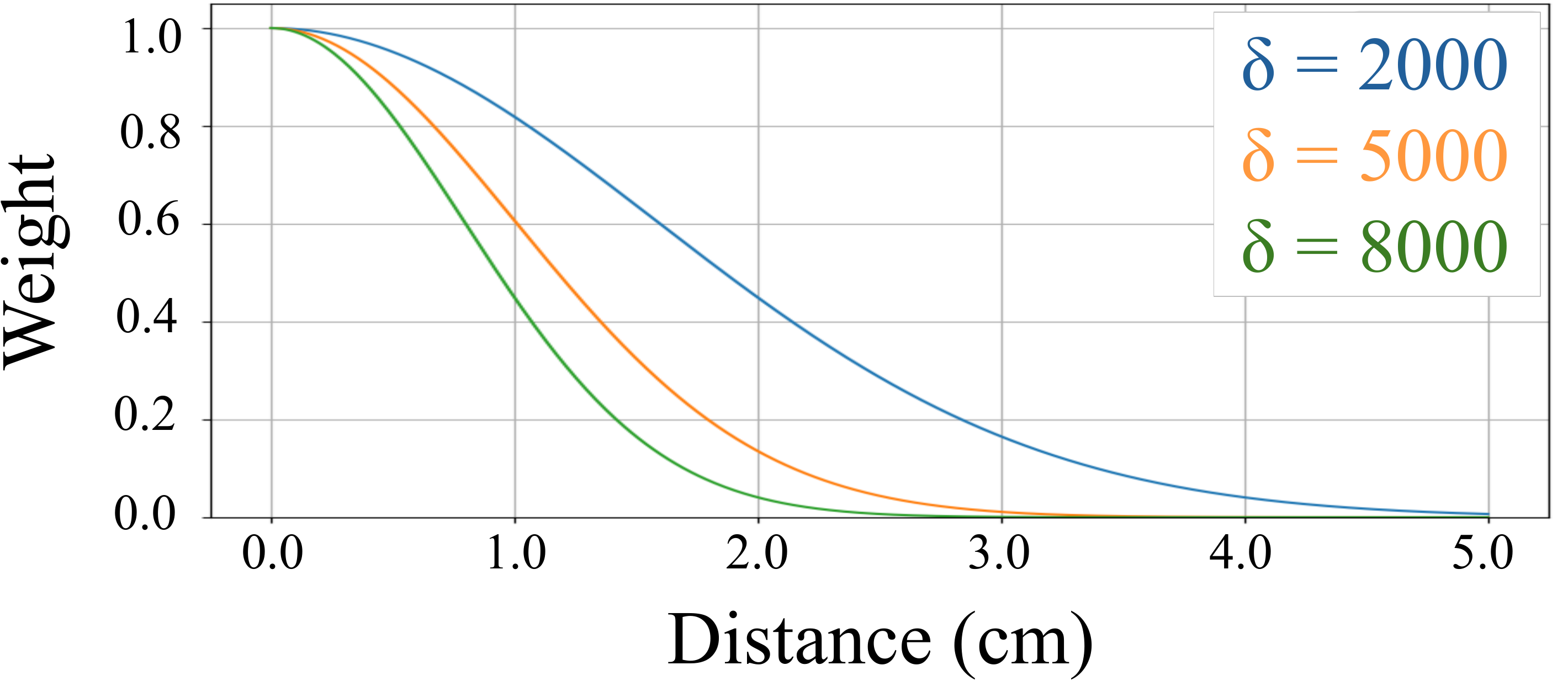}
    \caption{\textbf{Distance Profile per-vertex weight}: The relative importance $w_i$ of each hand vertex distance $\hat{\mathbf{D}}^i$ falls off exponentially depending on a sensitivity parameter $\delta$. For our chosen value $\delta$ = 5000, the weight virtually drops to 0 for distances further than 3 cm.}
    \label{fig:falloff_curve}
\end{figure}

\subsection{Grasp Energy-based Optimization}
\label{sec:method_optimization}
One of our main contributions lies in the refinement of initial candidate hand pose $\mathbf{H}_{\text{init}}$ (computed from $\hat{\boldsymbol{\gamma}}$ and $\hat{\boldsymbol{\theta}}_{\text{6D}}$) with respect to the target object, using the generated Distance Profile $\hat{\mathbf{D}}$.
Here, we convert the pose $\hat{\boldsymbol{\theta}}_{\text{6D}}$ from 6D to axis-angle representation $\hat{\boldsymbol{\theta}}$ for optimization.

We define the optimization objective as follows:
\begin{equation}
\label{eq:optimization_objective}
\mathcal{L}_{\text{opt}} =
\mathcal{L}_{\text{ge}} +
\gamma_{\text{root}} \mathcal{L}_{\text{root}} +
\gamma_{\text{reg}} \mathcal{L}_{\text{reg}} +
\gamma_{\text{col}} \mathcal{L}_{\text{col}},
\end{equation}
where $\gamma_{\text{root}}$, $\gamma_{\text{reg}}$ and $\gamma_{\text{col}}$ are weighting scales. 
%
%

\subsubsection{Distance-aware Grasp Energy Loss $\mathcal{L}_{\text{ge}}$.}
%
%
%
We first recompute the Distance Profile $\mathbf{D}_{\text{opt}}$ from the refined hand poses $\mathbf{H}_{\text{opt}}$ at each optimization step, using the same pre-sampled hand vertices and distance computation described in Sec.~\ref{subsec:representation}.
Next, we will align the refined configuration $\mathbf{D}_{\text{opt}}$ with the generated interaction profile $\hat{\mathbf{D}}$.
Here, since grasp functionality is primarily governed by near-contact geometry—where small spatial deviations can significantly alter contact locations and force exchange—we emphasize hand vertices that lie close to the object surface.
Specifically, we assign each hand vertex $i$ a distance-aware weight based on the predicted target distance $\hat{\mathbf{D}}^i$:
\begin{equation}
w_i = \exp\left( - \delta(\hat{\mathbf{D}}^i)^2 \right),
\end{equation}
where $\delta$ controls the fall-off sensitivity.
This formulation assigns larger weights to vertices with smaller target distances, prioritizing alignment in interaction-critical regions while reducing the influence of distant, non-interacting vertices, as shown in Fig.~\ref{fig:falloff_curve}.

The grasp energy loss is defined as:
\begin{equation}
\mathcal{L}_{\text{ge}} =
\frac{
\sum_{i=1}^{k}
w_i
\left(
\mathbf{D}_{\text{opt}}^i - \hat{\mathbf{D}}^i
\right)^2
}{
\sum_{i=1}^{k} w_i + \epsilon
}.
\end{equation}
where $\epsilon = 10^{-9}$ ensures numerical stability.
This term aligns the optimized grasp with the generated interaction profile, emphasizing geometrically and geometrically meaningful near-contact regions.

\subsubsection{Root Distance Loss $\mathcal{L}_{\text{root}}$.}

%
%
Let $r_{\text{opt}}$ denote the distance between the optimized root joint and its nearest object vertex. We preserve the intended hand–object proximity by:
\begin{equation}
\mathcal{L}_{\text{root}} =
\left\| r_{\text{opt}} - \hat{r} \right\|_2^2.
\end{equation}

\subsubsection{Rotation Regularization Loss $\mathcal{L}_{\text{reg}}$.}
%
%
To prevent excessive deviation from the diffusion prior and maintain anatomically plausible grasp, we regularize local joint rotations toward the initialization: 
\begin{equation}
\mathcal{L}_{\text{reg}} =
\sum_{j=1}^{J}
\left\|
\boldsymbol{\theta}_{\text{opt}}^{j} -
\hat{\boldsymbol{\theta}}^{j}
\right\|_2^2.
\end{equation}
This term constrains local pose variations while maintaining flexibility to satisfy the interaction constraints $\mathcal{L}_{\text{dp}}$ and $\mathcal{L}_{\text{root}}$.

\subsubsection{Collision Loss $\mathcal{L}_{\text{col}}$.}
%
%
We penalize interpenetration between the optimized hand and the object surface to ensure geometrically plausible hand–object interactions in those regions more weakly affected by the grasp energy loss.
Specifically, we compute signed penetration depths using surface normals and nearest-neighbor correspondences.
Note that the nearest-neighbor search is performed in every step, but the correspondence operation itself is non-differentiable.
Based on nearest-neighbor signed distances computed from each hand vertex $\mathbf{H}_{\text{opt}}^j$  and object point cloud $\mathbf{O}^i$, we form an object-to-hand $\mathbf{V}_{\text{o} \rightarrow \text{h}}^j$ and a hand-to-object vector $\mathbf{V}_{\text{h} \rightarrow \text{o}}^i$.
Signed penetration depths are computed with normals as:
\begin{equation}
d_{\text{h} \rightarrow \text{o}}^i
=
\langle \mathbf{N}_{\text{hand}}^{(i)}, \mathbf{V}_{\text{h} \rightarrow \text{o}}^i \rangle,
\qquad
d_{\text{o} \rightarrow \text{h}}^j
=
\langle \mathbf{N}_{\text{obj}}^{(j)}, \mathbf{V}_{\text{o} \rightarrow \text{h}}^j \rangle.
\end{equation}
Negative values indicate penetration.
We allow limited penetration up to a tolerance $\tau_{\text{col}} \ge 0$ and penalize only the excess penetration:
\begin{equation}
\mathcal{L}_{\text{col}} =
\sum_{i=1}^{P}
\max\big(-d_{\text{h} \rightarrow \text{o}}^i - \tau_{\text{col}}, 0\big)^2
+
\sum_{j=1}^{S}
\max\big(-d_{\text{o} \rightarrow \text{h}}^j - \tau_{\text{col}}, 0\big)^2.
\end{equation}

Overall, the refinement step enforces consistency with the generated Distance Profile and conditional signals while remaining close to the diffusion-initialized grasp prior, leading to geometrically coherent and semantically faithful hand–object interactions.


%
%
\section{Experiments}
\label{sec:experiments}

\subsection{Experiment Details}

\subsubsection{Datasets and Implementation.}
%
%
Existing HOI datasets cover a range of object categories but provide limited variation in hand identities. To address this limitation, we collect an internal real-world dataset, \textit{GraspShape}, comprising 350k frames from 100 distinct identities performing diverse grasps with 5 objects.
In addition, we utilize a subset \footnote{Due to license restrictions, we can use 25 of the 51 available GRAB objects. Please refer to the supplementary material for the object list.} of the GRAB dataset~\cite{grab}, which includes 25 objects and 3 identities. 
%
%
%
See the supplement for more details of the implementation and model architecture.

\begin{table*}[t]
\centering
\caption{\textbf{Quantitative results on the scaled objects with various hand directions and hand shapes.} \textbf{Init}: our initial prediction, \textbf{GE-opt}: Distance Profile optimization. The root distance is set to 7 mm. The standard deviations over generated samples are shown in parentheses.
}
\scalebox{0.9}{
\begin{tabular}{llrrrrrr}
\hline
\noalign{\smallskip}
 \multicolumn{8}{c}{GraspShape Objects (70, 80, 90, 110, and 120\% scales)}  \\
\noalign{\smallskip}
\hline
\noalign{\smallskip}
 \,  & Scale \, & \, Pen [\%] $\downarrow$ \,  &  \, F-Con [\%] $\uparrow$ \, &  \, V-Con [\%] $-$ \, &  \, $r$-Dist [mm] $\downarrow$ \, & Angle [degree] $\downarrow$ &  \, Div [mm] $\uparrow$ \, \\ 
\noalign{\smallskip}
\hline
\noalign{\smallskip}
Baseline  \,& All \,  & 65.89 (39.62) \, & 99.57 (3.03) \, & 23.55 (9.69) \, & 11.99 (6.50) \, & \textbf{3.95} (2.06) \,   &  18.31 (6.48) \,\\
Ours (Init) \,& All \,  & 22.60 (16.73) \, & 98.16 (7.19) \, & 23.78 (10.19) \, & 10.13 (5.78) \, & 5.49 (2.29) \,   & \textbf{20.43} (6.80) \,\\
Ours (GE-opt) \, & All \,  & \textbf{5.52} (3.86) \, & \textbf{100.00} (0.00) \, & 31.81 (9.12) \, & \textbf{5.74} (5.60) \, & 6.02 (3.11) \,  & 20.23 (6.77) \, \\
\noalign{\smallskip}
\noalign{\smallskip}
Baseline  \,& 70\% \,  & 84.38 (27.81) \, &  99.48 (4.13) \, & 9.16 (4.57) \, & 21.87 (7.90) \, & \textbf{3.54} (1.85)  \,   & 14.95 (5.93) \,\\
Ours (Init) \,& 70\% \,  & 15.10  (13.43) \, & 93.75 (21.52) \, & 7.26 (5.51) \,  & 17.17 (7.10) \, & 5.47 (2.29) \, & \textbf{20.53} (6.89)\,  \\
Ours (GE-opt) \, & 70\% \,  & \textbf{4.17} (4.62) \, & \textbf{100.00} (0.00)  \, & 28.76 (8.48) \, & \textbf{6.07} (5.25) \, & 5.67 (3.00) \,   & 19.70 (6.72) \,  \\
\noalign{\smallskip}
\noalign{\smallskip}
Baseline  \,& 120\% \,  &  65.10 (45.03) \, & \textbf{100.00} (0.00)  \, & 32.44 (12.80) \, & 10.97 (7.99)  \, & \textbf{4.18} (2.17)  \,   & 18.20 (6.55) \,\\
Ours (Init) \, & 120\% \,  & 39.58 (31.37) \, & \textbf{100.00} (0.00) \, & 43.91 (14.75) \, & 11.84 (6.44)  \, & 5.47 (2.29)  \, & 20.53 (6.89) \, \\
Ours (GE-opt) \,  & 120\% \,  & \textbf{8.33}  (3.51) \, & \textbf{100.00} (0.00) \, & 34.37 (9.89)  \,  & \textbf{6.62} (6.58) \, & 6.14 (3.10) \,  & \textbf{20.86} (6.91) \, \\
\noalign{\smallskip}
\hline
\noalign{\smallskip}
\multicolumn{8}{c}{GRAB Objects (70, 80, 90, 100, 110 and 120\% scales)}  \\
\noalign{\smallskip}
\hline
\noalign{\smallskip}
 \,   & Scale \, & \, Pen [\%] $\downarrow$ \,  &  \, F-Con [\%] $\uparrow$ \, &  \, V-Con [\%] $-$ \, &  \, $r$-Dist [mm] $\downarrow$ \, & Angle [degree] $\downarrow$ &  \, Div [mm] $\uparrow$ \,  \\ 
\noalign{\smallskip}
\hline
\noalign{\smallskip}
Baseline  \,& All \,  & 76.88 (33.15) \, & 97.09 (2.72) \, & 10.34 (4.89) \, & 24.71 (8.61) \, &  4.07 (2.21) \,   & 14.02 (6.77) \,\\
Ours (Init) \,  & All \,  & 54.69 (29.70)  \, & 99.74 (1.66) \, & 21.21 (7.38) \, & 9.38 (6.03) \, & \textbf{3.83} (1.67) \,  & 14.01 (5.02) \,  \\
Ours (GE-opt) \,   & All \,  & \textbf{0.00} (0.00) \, & \textbf{100.00} (0.00) \, & 24.27 (6.53) \, & \textbf{6.48} (4.83) \, & 4.07 (2.13) \, & \textbf{14.11} (4.78) \,  \\
\noalign{\smallskip}
\noalign{\smallskip}
Baseline  \,& 70\% \,  &  71.88 (25.13) \, & 97.92 (8.07) \, & 4.77 (1.93) \, & 30.23 (6.84) \, & 4.01 (2.00) \,   & \textbf{14.24} (6.30) \,\\
Ours (Init) \, & 70\% \,  &  11.98 (19.97) \, & 98.44 (9.93) \, & 6.59 (3.61) \,  & 10.02 (6.56) \, & \textbf{3.83} (1.67) \, & 14.02 (5.02) \, \\
Ours (GE-opt) \,  & 70\% \,  & \textbf{0.00} (0.00) \, & \textbf{100.00} (0.00) \, & 20.56 (6.08) \, & \textbf{5.95} (4.73)  \, & 4.23 (2.26) \, & 13.72 (4.63)  \,  \\
\noalign{\smallskip}
\noalign{\smallskip}
Baseline  \,& 120\% \,  & 62.45 (35.23) \, & \textbf{100.00} (0.00)  \, & 14.54 (7.32)  \, & 16.22 (8.23) \, & 4.39 (2.02)  \,   & 14.17 (6.44) \,\\
Ours (Init) \, & 120\% \,  & 97.92 (11.18) \, & \textbf{100.00} (0.00) \, & 36.03 (9.55) \, & 12.39 (7.39)  \, & \textbf{3.83} (1.67)  \, & 14.01 (5.02)  \,  \\
Ours (GE-opt) \,  & 120\% \,  & \textbf{0.00} (0.00) \, & \textbf{100.00} (0.00) \, & 26.83 (6.56)  \, & \textbf{8.48} (5.52) \, & 4.13 (2.14) \, & \textbf{14.62} (4.92) \,  \\
\noalign{\smallskip}
\hline
\end{tabular}
}
\label{table:quantitative_result_scaled_obj}
\end{table*}

\begin{table*}[t]
\centering
\caption{\textbf{Object-wise evaluations with various hand directions and hand shapes.}  \textbf{Init}: our initial prediction, \textbf{GE-opt}: grasp energy-based optimization. The root distance is set to 7 mm. The standard deviations over generated samples are shown in parentheses.
}
\scalebox{0.9}{
\begin{tabular}{llrrrrrr}
\hline
\noalign{\smallskip}
 \multicolumn{8}{c}{GraspShape Objects (70, 80, 90, 110, and 120\% scales)}  \\
\noalign{\smallskip}
\hline
\noalign{\smallskip}
 \, & Object \,  & \, Pen [\%] $\downarrow$ \,  &  \, F-Con [\%] $\uparrow$ \, &  \, V-Con [\%] $-$ \, &  \, $r$-Dist [mm] $\downarrow$ \, & Angle [degree] $\downarrow$ &  \, Div [mm] $\uparrow$ \, \\ 
\noalign{\smallskip}
\hline
\noalign{\smallskip}
Baseline  \,& Mug \,  & 43.75 (44.33) \, & 99.74 (2,07) \, & 28.32 (11.91) \, & 12.24 (7.93)  \, & \textbf{3.95} (1.92)  \,   & 18.81 (6.79)  \,\\
Ours (Init) \, & Mug \,  & 22.19 (23.41) \, &  94.69 (18.50) \, & 22.85 (11.26)  \, & 10.23 (5.58)  \, & 5.06 (2.39)  \,   & 22.13 (8.21)  \, \\
Ours (GE-opt) \, & Mug \,  &  \textbf{0.63} (0.89) \, & \textbf{100.00} (0.00) \, & 32.95 (9.76) \, & \textbf{4.90} (4.10) \, & 5.67 (2.73) \,  & \textbf{21.60} (8.10) \, \\
\noalign{\smallskip}
\noalign{\smallskip}
Baseline  \,& Can \,  & 80.73 (30.70) \, & \textbf{100.00} (0.00)  \, & 23.24 (7.65) \, & 16.33 (6.50)  \, & \textbf{3.77} (2.00)  \,   & 16.64 (6.02)  \,\\
Ours (Init)  \, & Can \,   & 36.6  (23.88) \, & 99.48 (2.90) \, &  27.98 (9.71) \, & 8.67 (5.23) \, & 5.02 (2.14) \,  &  \textbf{18.70} (5.70)  \\
Ours (GE-opt) \, & Can \,    &  \textbf{3.13} (2.77) \, & \textbf{100.00} (0.00) \, & 35.28 (8.63) \,  & \textbf{5.79} (6.15)  \, & 5.12 (2.84) \,  & 18.42 (5.67) \,\\
\noalign{\smallskip}
\noalign{\smallskip}
Baseline   \,& Smartphone \,  & 73.18 (43.83) \, & 98.96 (7.03)  \, & 19.11 (9.50) \, & 7.40 (5.07) \, & \textbf{4.14} (2.25)  \,   & 19.45 (6.63)  \,\\
Ours (Init) \, & Smartphone \,  & 30.63 (5.37) \, & 99.74 (2.07) \, & 20.36 (9.77) \, & 11.51 (6.50) \, & 6.33 (2.35) \,  & 20.76 (6.76) \,  \\
Ours (GE-opt) \, & Smartphone \,   & \textbf{12.81} (1.92) \, & \textbf{100.00} (0.00) \, & 27.41 (9.09)  \, & \textbf{6.40} (6.31) \, & 7.23 (3.69) \,  & \textbf{20.91} (6.76)  \, \\
\noalign{\smallskip}
\hline
\hline
\noalign{\smallskip}
\multicolumn{8}{c}{GRAB Objects (70, 80, 90, 100, 110 and 120\% scales)}  \\
\noalign{\smallskip}
\hline
\noalign{\smallskip}
 \, & Object \,  & \, Pen [\%] $\downarrow$ \,  &  \, F-Con [\%] $\uparrow$ \, &  \, V-Con [\%] $-$ \, &  \, $r$-Dist [mm] $\downarrow$ \, & Angle [degree] $\downarrow$ &  \, Div [mm] $\uparrow$ \,  \\ 
\noalign{\smallskip}
\hline
\noalign{\smallskip}
Baseline  \,& Apple \,  & 77.04 (26.28)  \, & 99.20 (3.75) \, & 10.72 (4.67) \, & 25.27 (7.44)  \, & 3.71 (1.98) \,   & \textbf{14.48} (5.89)  \,\\
Ours (Init) \, & Apple \,  & 52.86 (24.97) \, & 99.74 (2.07)  \, & 28.01 (9.10)  \, & 8.81 (5.15)  \, & 3.27 (1.67) \, & 13.32 (4.37) \,  \\
Ours (GE-opt) \, & Apple \,  & \textbf{0.00} (0.00) \, & \textbf{100.00} (0.00) \, & 34.22 (6.79) \, & \textbf{5.18} (3.19) \, & \textbf{3.13} (1.67) \, & 12.70 (4.27) \,  \\
\noalign{\smallskip}
\noalign{\smallskip}
Baseline  \,& Doorknob \,  & 73.30 (40.01) \, & \textbf{100.00} (0.00) \, & 13.29 (5.48) \, & 21.71 (9.85) \, & \textbf{3.14} (1.57)  \,   & 12.81 (5.00)  \,\\
Ours (Init) \, & Doorknob \,  & 71.61 (37.06) \, & \textbf{100.00} (0.00) \, & 19.28 (6.78) \, & 10.56 (7.59) \, &  4.89 (1.99)\, & 13.05 (5.29) \,  \\
Ours (GE-opt) \, & Doorknob \,    & \textbf{0.00} (0.00) \, & \textbf{100.00} (0.00) \, & 18.91 (6.49) \, & \textbf{7.87} (6.64) \, & 5.34 (2.64) \, & \textbf{13.48} (4.98) \,  \\
\noalign{\smallskip}
\noalign{\smallskip}
Baseline  \,& Torus \,  & 74.24 (37.88) \, &  99.28 (3.77) \, & 10.52 (4.76)  \, &  24.38 (8.44) \, & 3.97 (1.96)   \,   & 14.47 (6.34)  \,\\
Ours (Init) \, & Torus \,   & 39.58 (27.06) \, & 99.48 (2.90) \, & 16.34 (6.26) \,  & 8.76 (5.34) \, & \textbf{3.33} (1.34) \,  & 15.67 (5.42) \, \\
Ours (GE-opt)\, & Torus \,    & \textbf{0.00} (0.00) \, & \textbf{100.00} (0.00) \, & 19.67 (6.31) \, & \textbf{6.40} (4.66) \, & 3.75 (2.08) \,  & \textbf{16.17} (5.09) \,  \\
\noalign{\smallskip}
\hline
\end{tabular}
}
\label{table:quantitative_result_scaled_obj_category}
\end{table*}

\begin{figure*}[tb]
  \centering
  \includegraphics[width=0.84\linewidth]{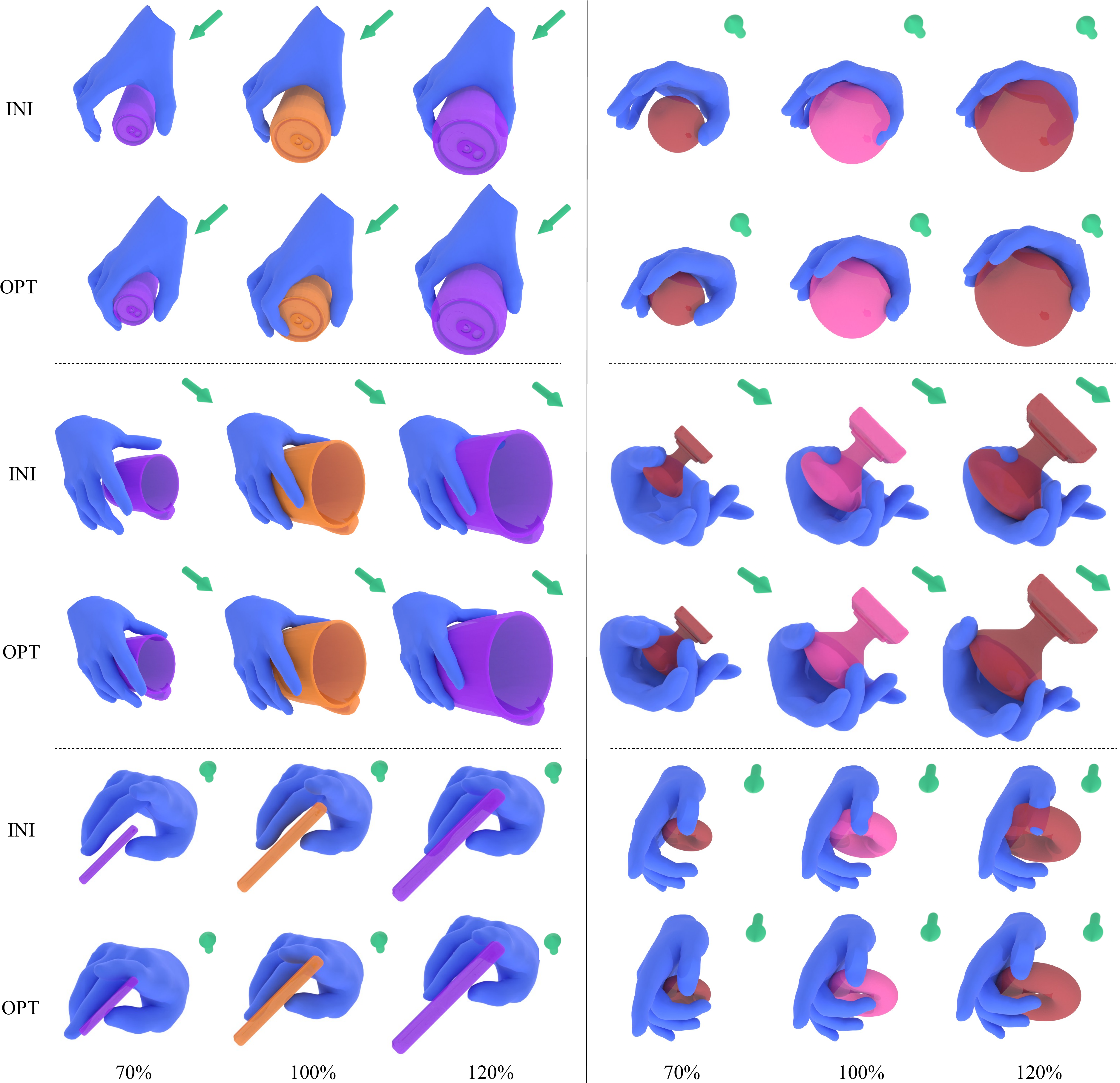}
  \caption{\textbf{Qualitative results on scaled objects with various hand directions and hand shapes.} \textbf{Left}: GraspShape objects, \textbf{Right}: GRAB objects. \textbf{INI}: Initial pose, \textbf{OPT}: Optimized pose. The optimization step refines the initial poses into physically plausible and functional grasp poses.
  }
  \label{fig:qualitative_results_scaled_obj}
\end{figure*}

\begin{figure*}[t]
\centering
\begin{minipage}{0.49\textwidth}
    \centering
      \includegraphics[width=0.7\linewidth]{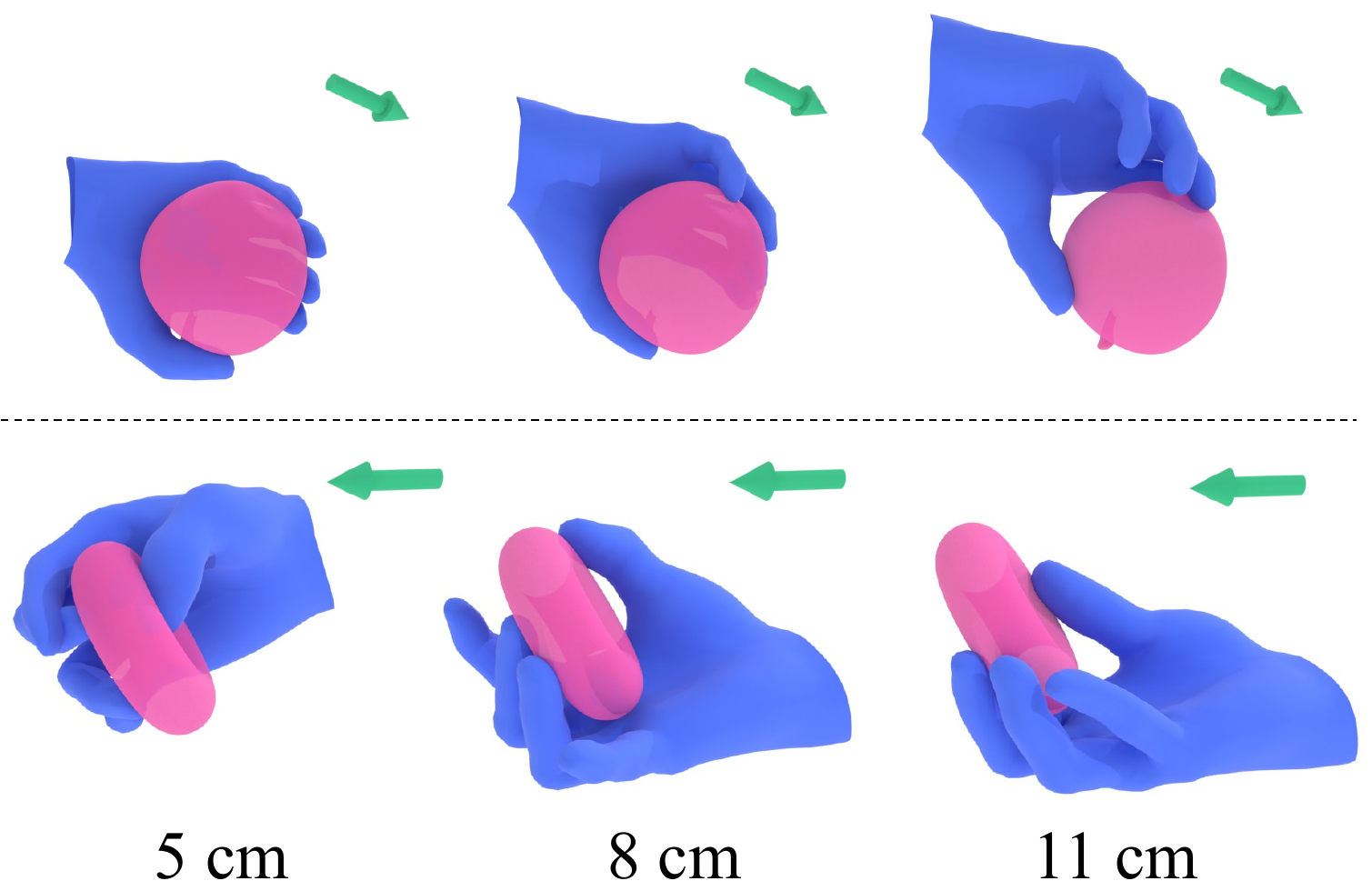}
      \vspace{-2mm}
      \caption{\textbf{Qualitative results with varying root distance conditioning}. This signal effectively controls the closeness of the interaction while preserving plausibility.
      }
      \label{fig:ablation_root_dist}
      \vspace{-3mm}
\end{minipage}
\hfill
\begin{minipage}{0.49\textwidth}
    \centering
      \includegraphics[width=0.8\linewidth]{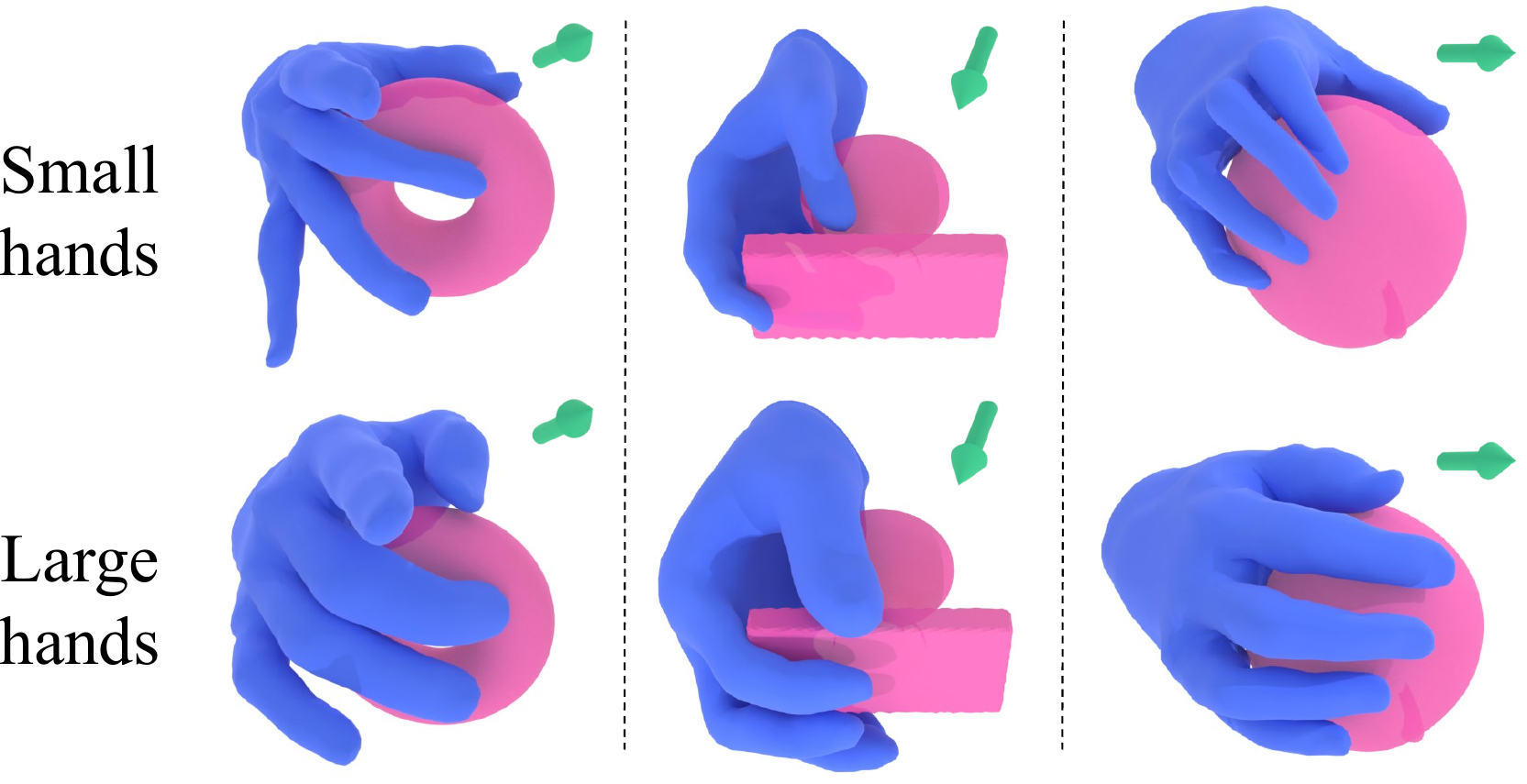}
      \caption{\textbf{Qualitative results with extreme hand shape conditioning}. The semantics of the interaction are effectively maintained for different hand identities.
      }
      \label{fig:ablation_hand_shape}
      \vspace{-3mm}
\end{minipage}
\vspace{-4mm}
\end{figure*}

\subsubsection{Evaluation Protocol.}
No prior work addresses the task of balancing geometrically plausible grasp generation while maintaining fidelity to the multi-modal deterministic controllable inputs with dedicated mechanisms.
Therefore, we establish a strong baseline based on the state-of-the-art Diffusion Transformer (DiT) architecture (same as our full model), which accepts a series of controllable signals and generates only translation and rotations (\ie without predicting a Distance Profile).
We also provide the evaluation for the initial prediction from our system.


To evaluate the system's sensitivity to various object shapes and scales, we use six objects under scaled-variation settings: 3 training-category objects from GraspShape (mug, can, and smartphone) with 70, 80, 90, 110, and 120\% scales, and 3 testing-category objects from GRAB (apple, doorknob, and torus) with 70 $\sim$ 120\% scales.
Furthermore, we provide more results with additional scaled objects and hand shapes in the supplementary material.
Note that none of these objects are used during training; they are reserved exclusively for evaluation to assess generalization, and training objects are not scaled (\ie 100\%).
This evaluation protocol creates a wider variety of unseen objects and scales for an evaluation of our system trained only on single-scale objects. 
In addition, we consider diverse conditional settings by sampling controllable signals from a normal distribution.
For hand shape parameters, we further sample from a uniform distribution within $[-3.0, 3.0]$ to evaluate performance under extreme shape variations.

For each object and each scale, we generate 64 samples (\ie 384 samples per object group across 70--120\% scales) and follow existing works~\cite{zhang2025bimart, paschalidis2025cwgrasp} to report common metrics to evaluate geometrical plausibility with their standard deviations over generated samples (see parentheses in each table).
\begin{itemize}
    \item \textbf{Pen}: Measures geometrical plausibility, defined as the percentage of grasp frames exhibiting hand–object interpenetration, using a 1 cm distance threshold.
    \item \textbf{F-Con}: Evaluates frame-level contact fidelity, computed as the percentage of frames in which the hand establishes contact with the object within a 1 cm threshold.
    \item \textbf{V-Con}: Measures vertex-level contact fidelity, defined as the percentage of hand vertices that are in contact with the object within a 1 cm threshold. 
    Note that V-Con also includes vertices on the outer palm, which typically do not contact the object; therefore, values around 50\% represent the practical upper bound. 
    Moreover, there is no single optimal value for this metric since valid grasps may involve different contact patterns. 
    Nonetheless, this metric is still valuable to evaluate the controllability aspect.
    \item \textbf{$r$-Dist}: Assesses the hand-object proximity, computed as the mean Euclidean error between the root distances from the input signals and those of the generated grasps.
    \item \textbf{Angle}: Evaluates directional consistency, defined as the mean angular error between the direction vectors from the input signals and those of the generated grasps.
\end{itemize}
Furthermore, although our primary focus is on ensuring the geometrical plausibility of generated grasps under diverse controllable signals—which may constrain the outputs to specific regions of the training distribution—we additionally report a diversity metric, "Div", for completeness. 
This metric quantifies generative diversity as the mean pairwise vertex distance across all generated grasps, averaged over all grasp pairs, after rigidly aligning the hands to the same wrist position and palm orientation, essentially indicating the diversity of fingers given a series of controllable signals.

\begin{table*}[t]
\parbox{.49\linewidth}{
\centering
\caption{\textbf{Quantitative results with varying root distances.}}
\scalebox{1.0}{
            \begin{tabular}{lrrrr}
            \hline
            \noalign{\smallskip}
            \multicolumn{1}{l}{Metrics} & \, Pen [\%] $\downarrow$ \,  &  \, F-Con [\%] $\uparrow$ \, &  \, V-Con [\%] $-$ \, \\ 
            \noalign{\smallskip}
            \hline
            \noalign{\smallskip}
            5 cm & 0.00 (0.00) \, & 100.00 (0.00) \, & 24.40 (7.99) \,  \\
            8 cm & 0.52 (4.13) \, & 100.00 (0.00) \, & 26.23 (5.39) \,  \\
            11 cm & 0.00 (0.00) \, & 100.00 (0.00) \, & 17.64  (4.68) \,  \\
            \noalign{\smallskip}
            \hline
            \hline
            \multicolumn{1}{l}{Metrics} &  \, $r$-Dist [mm] $\downarrow$ \, & Angle [degree] $\downarrow$ \, &  \, Div [mm] $\uparrow$ \,  \\ 
            \noalign{\smallskip}
            \hline
            \noalign{\smallskip}
            5 cm & 15.18 (5.99) \, & 7.47 (3.01) \, & 14.23 (5.71) \, \\
            8 cm & 7.25 (3.81) \, & 3.97 (2.18) \, & 13.25 (5.10) \,  \\
            11 cm & 8.33 (4.69) \, & 4.47  (2.25) \, & 14.22 (5.38) \,  \\
            \noalign{\smallskip}
            \hline
            \label{table:root_dist}
            \end{tabular}
}
}
\hfill
\parbox{.49\linewidth}{
\centering
\caption{\textbf{Quantitative results with extreme hand shapes.}}
\scalebox{1.0}{
            \begin{tabular}{lrrrr}
            \hline
            \noalign{\smallskip}
            \multicolumn{1}{l}{Metrics} & \, Pen [\%] $\downarrow$ \,  &  \, F-Con [\%] $\uparrow$ \, &  \, V-Con [\%] $-$ \, \\ 
            \noalign{\smallskip}
            \hline
            \noalign{\smallskip}
            Normal & 0.00 (0.00) \, & 100.00 (0.00) \, & 28.09  (5.93) \,  \\
            Uniform & 0.00 (0.00) \, & 100.00 (0.00) \, & 26.03 (6.75) \,  \\
            \noalign{\smallskip}
            \hline
            \hline
            \multicolumn{1}{l}{Metrics}  &  \, $r$-Dist [mm] $\downarrow$ \, & Angle [degree] $\downarrow$ \, &  \, Div [mm] $\uparrow$ \,  \\ 
            \noalign{\smallskip}
            \hline
            \noalign{\smallskip}
            Normal & 6.67 (4.24) \, & 3.72 (2.08) \, & 12.41 (4.65) \,  \\
            Uniform & 7.25 (5.38) \, & 5.80 (3.23) \, & 16.04 (6.56) \,  \\
            \noalign{\smallskip}
            \hline
            \label{table:hand_shape}
            \end{tabular}
}
}
\end{table*}

\subsection{Results}
\label{subsec:results}
%
%

%

\subsubsection{Contact Fidelity}
Table~\ref{table:quantitative_result_scaled_obj} shows that DCGrasp achieves significantly low penetration rates (Pen), indicating that the framework effectively enforces penetration-robust hand–object interactions.
Even under substantial scale variations (70$\sim$120\%), penetration remains minimal — often nearly zero — highlighting strong robustness to geometric transformations.
Furthermore, the refinement process attains high frame-level contact (F-Con, nearly 100\%) and stable vertex-level contact (V-Con), suggesting that generated grasps establish meaningful and spatially coherent hand–object interactions.
DCGrasp preserves consistent contact across scales, validating the effectiveness of the grasp energy-based approach.
Notably, even our initial prediction provides a better penetration ratio compared to the baseline, \eg 22.60\% vs. 65.89\% on average with GraspShape, while maintaining on-par results on the contact ratio (F-Con), \eg 98.16\% vs. 99.57\% on average with GraspShape.
This result suggests that the Distance Profile representation itself may also help the model learn the spatial relationship of hand-object interaction.

Table~\ref{table:quantitative_result_scaled_obj_category} shows that DCGrasp consistently generates grasps with highly geometrical plausibility across different objects, achieving 0\% Pen, 100\% F-Con, and approximately 20\% V-Con on average.
Note that the geometrical errors on smartphones are slightly higher than other objects.
This is due to the initial predictions that contain severe interpenetration. During optimization, the collision penalty may pull the hand in different directions, occasionally leading to local minima. See the supplement for more discussion on failure cases.

Fig.~\ref{fig:qualitative_results_scaled_obj} visualizes both the initial and optimized grasp poses.
The grasp energy–based optimization naturally guides the fingers to remain close to the object surface while avoiding excessive interpenetration.
Notably, many initial poses on the 70\% scaled objects already satisfy frame-level contact and thus trigger F-Con (\eg more than 90\%), as commonly reported in prior works; however, these poses do not always correspond to functional grasps as shown in the visualizations.
DCGrasp effectively refines such superficial contacts into stable and semantically valid grasping poses.

\subsubsection{Controllability}
DCGrasp shows accurate adherence to the control signals.
The root distance error ($r$-Dist) remains consistently low across all settings, \eg 5.74 mm on GraspShape and 6.48 mm on GRAB objects,
indicating that global hand–object proximity can be reliably regulated by our system.
Similarly, the angular deviation (Angle) between input and generated grasps is small, \eg 6.02 and 4.07 degrees on the GraspShape and GRAB objects, respectively, demonstrating precise directional controllability.

Furthermore, Fig.~\ref{fig:qualitative_results_hdv} visualizes poses generated by DCGrasp under closely varying directional inputs.
The resulting grasps faithfully follow the specified direction trajectories while maintaining natural hand configurations.
These results highlight that DCGrasp supports flexible control, making it suitable for a wide range of applications requiring explicit grasp manipulation, as discussed in Sec.~\ref{sec:intro}.

\subsubsection{Diversity}
Our full pipeline preserves diversity levels comparable to those of the initial predictions and the baseline outputs.
While the controllable signals guide the diffusion sampling trajectory toward specific regions of the learned distribution, our system maintains generative variability.
On GraspShape, the diversity measures 20.43 mm for raw samples from the diffusion prior and 20.23 mm after optimization.
Together with the results discussed above, these findings show that the grasp energy-based refinement process preserves generative diversity while improving geometrical plausibility and generalizing effectively to novel object shapes and scales.

\subsubsection{Varying Root Distance Conditioning}
We further evaluate explicit control over root distance using three distinct target distances (5 cm, 8 cm, 11 cm), as shown in Table~\ref{table:root_dist}.
Importantly, increasing the target root distance does not degrade frame-level contact stability. 
Instead, it primarily modulates the vertex-level contact ratio, effectively controlling the degree of surface engagement. 
When smaller root distances are specified (\eg 5 cm), the model balances hand-object proximity with geometrical plausibility, maintaining low penetration and stable contact while approximating the desired distance.
Qualitative results in Fig.~\ref{fig:ablation_root_dist} show smooth and continuous variation in hand–object proximity.
The transition across different distance settings is geometrically consistent, demonstrating that root distance control is both interpretable and stable.

\subsubsection{Extreme Hand Shape Conditioning}
To assess robustness to shape variations, we test DCGrasp with both normally distributed and uniformly sampled extreme hand shape parameters within $[-3.0, 3.0]$.
As reported in Table~\ref{table:hand_shape}, DCGrasp maintains low penetration (0\% on Pen) and stable contact (100\% on F-Con and 26\% on V-Con) under extreme shape conditions. 
Fig.~\ref{fig:ablation_hand_shape} shows that the synthesized grasps adapt naturally to significantly different hand geometries, including slender and wide palms.
Despite large identity variations, interaction semantics remain consistent across shapes.
This demonstrates that the proposed grasp energy-based approach generalizes grasp structure beyond specific hand identities, enabling articulation and contact formation to remain stable under extreme morphological changes.

%
%
\section{Discussion and Future Work}
\label{sec:discussion}
%


%
Although DCGrasp is based on a parametric hand model, the proposed grasp energy term is conceptually general.
It simply encodes grasp interaction as weighted signed distances between two surfaces and does not inherently depend on a specific kinematic structure or embodiment.
This abstraction suggests that the grasp energy term can be naturally extended to other articulated effectors, such as multi-fingered robotic hands or even the full human body with different kinematic chains.
By redefining the sampled surface vertices on the manipulator and computing distances to target objects, the same interaction representation could be preserved.
This opens the possibility of transferring grasp semantics learned from human demonstrations to robotic platforms.
Beyond hands, the formulation may also be applicable to full-body interactions (\eg human– or robot–object contact involving arms, legs, or torso), where modeling spatial proximity between articulated surfaces is essential.
Extending the grasp energy term to whole-body interaction modeling could provide a unified representation for diverse embodied agents.





%
%
\section{Conclusion}
\label{sec:conclusion}
We introduced DCGrasp, a multi-modal diffusion framework for distance-aware controllable hand–object grasp synthesis built upon a novel grasp energy term. 
By encoding per-vertex proximity between the hand and object as Distance Profile and weighting it based on the hand-object proximity, the proposed grasp energy term produces families of semantically similar grasps in near-contact regions while remaining invariant to object and hand identity. 
%
%
DCGrasp combines diffusion-based Distance Profile generation with grasp energy–guided optimization, effectively decoupling interaction modeling from object-specific geometry. 
This design not only enables flexible conditioning on multiple user-defined signals but also enhances robustness and generalization across diverse object geometries and scales. 
Our experiments show that DCGrasp produces high-quality, geometrically plausible grasps, even with unseen objects with varying shapes and scales.
Our work provides a promising foundation for future research in hand–object interaction modeling, with potential impact on robotics, AR/VR, and synthetic data generation.







\bibliographystyle{ACM-Reference-Format}
\bibliography{submission_base}

@String{Computing = "Computing" }

@String(PAMI 	=	{{Transactions on Pattern Analysis and Machine Intelligence (TPAMI)}})

@String(CVPR 	=	{{Computer Vision and Pattern Recognition (CVPR)}})

@String(ICCV 	=	{{International Conference on Computer Vision ({ICCV})}})

@String(ECCV 	=	{{European Conference on Computer Vision (ECCV)}})

@String(BMVC 	=	{{British Machine Vision Conference (BMVC)}})

@String(TOG 	=	{{Transactions on Graphics (TOG)}})

@String(ICLR 	=	{{International Conference on Learning Representations (ICLR)}})

@STRING(I3DV 	=	{{International Conference on 3D Vision (3DV)}})

@STRING(CGF 	=	{{Computer Graphics Forum (CGF)}})

@STRING(SIGG	=	{{International Conference on Computer Graphics and Interactive Techniques (SIGGRAPH)}})

@STRING(AAAI	=	{{AAAI Conference on Artificial Intelligence}})

@String(ICRA	= 	{{International Conference on Robotics and Automation (ICRA)}})

@String(IROS	= 	{{International Conference on Intelligent Robots and Systems (IROS)}})

@String(CORL	= 	{{Conference on Robot Learning (CoRL)}})

@String(RAL     = 	{{Robotics and Automation Letters (RA-L)}})

@ArtifactSoftware{R,
    title = {R: A Language and Environment for Statistical Computing},
    author = {{R Core Team}},
    organization = {R Foundation for Statistical Computing},
    address = {Vienna, Austria},
    year = {2019},
    url = {https://www.R-project.org/},
}

@inproceedings{contactdb,
    author      =   {Samarth Brahmbhatt and Cusuh Ham and Charles C. Kemp and James Hays},

    title       =   {{ContactDB}: {A}nalyzing and Predicting Grasp Contact via Thermal Imaging},
    booktitle   =   CVPR,
    year        =   {2019},
    pages       =   {8709-8719}
}

@inproceedings{grasping_field,
    author      =   {Korrawe Karunratanakul and Jinlong Yang and Yan Zhang and Michael J. Black and Krikamol Muandet and Siyu Tang},
    booktitle   =   I3DV, 
    title       =   {Grasping Field: Learning Implicit Representations for Human Grasps}, 
    year        =   {2020},
    pages       =   {333-344},
}

@article{mano,
    author       = {Javier Romero and Dimitrios Tzionas and Michael J. Black},
    title       =   {Embodied hands: {M}odeling and capturing hands and bodies together},
    year        =   2017,
    volume      =   {36},
    number      =   {6},
    pages       =   {1-17},
    journal     =   TOG
}

@inproceedings{hand_gaussians,
    author      =   {Srinath Sridhar and Antti Oulasvirta and Christian Theobalt},
    title       =   {Interactive Markerless Articulated Hand Motion Tracking Using RGB and Depth Data},
    booktitle   =   ICCV,
    year        =   {2013},
    pages       =   {2456-2463}
}

@inproceedings{contactgrasp,
    author      =   {Samarth Brahmbhatt and Ankur Handa and James Hays and Dieter Fox},
    booktitle   =   IROS, 
    title       =   {{ContactGrasp}: {F}unctional Multi-finger Grasp Synthesis from Contact}, 
    year        =   {2019},
    pages       =   {2386-2393},
}

@inproceedings{contactopt,
    author      =   {Patrick Grady and Chengcheng Tang and Christopher D. Twigg and Minh Vo and Samarth Brahmbhatt and Charles C. Kemp},
    title       =   {{ContactOpt}: {O}ptimizing Contact To Improve Grasps},
    booktitle   =   CVPR,
    year        =   {2021},
    pages       =   {1471-1481}
}

@inproceedings{grab,
    title       =   {{GRAB}: {A} Dataset of Whole-Body Human Grasping of Objects},
    author      =   {Omid Taheri and Nima Ghorbani and Michael J. Black and Dimitrios Tzionas},
    booktitle   =   ECCV,
    year        =   {2020},
    pages       =   {581-600}
}

@inproceedings{flex,
    author      =   {Purva Tendulkar and D{\'{\i}}dac Sur{\'{\i}}s and Carl Vondrick},
    title       =   {{FLEX}: {F}ull-Body Grasping Without Full-Body Grasps},
    booktitle   =   CVPR,
    year        =   {2023},
    pages       =   {21179-21189}
}

@inproceedings{goal,
    author      =   {Omid Taheri and Vasileios Choutas and Michael J. Black and Dimitrios Tzionas},
    title       =   {{GOAL}: {G}enerating {4D} Whole-Body Motion for Hand-Object Grasping},
    booktitle   =   CVPR,
    year        =   {2022},
    pages       =   {13263-13273}
}

@inproceedings{dgrasp,
    author      =   {Sammy Joe Christen and
                  Muhammed Kocabas and
                  Emre Aksan and
                  Jemin Hwangbo and
                  Jie Song and
                  Otmar Hilliges},
    title       =   {{D-Grasp}: {P}hysically Plausible Dynamic Grasp Synthesis for Hand-Object Interactions},
    booktitle   =   CVPR,
    year        =   {2022},
    pages       =   {20577-20586}
}

@article{manipnet,
    author       = {He Zhang and Yuting Ye and Takaaki Shiratori and Taku Komura},

    title       =   {{ManipNet}: {N}eural Manipulation Synthesis with a Hand-Object Spatial Representation},
    year        =   {2021},
    volume      =   {40},
    number      =   {4},
    pages       =  {121:1--121:14},
    journal     =   TOG,
}

@inproceedings{ganhand,
    author      =   {Enric Corona and
                  Albert Pumarola and
                  Guillem Aleny{\`{a}} and
                  Francesc Moreno{-}Noguer and
                  Gr{\'{e}}gory Rogez},
    title       =   {{GanHand}: {P}redicting Human Grasp Affordances in Multi-Object Scenes},
    booktitle   =   CVPR,
    year        =   {2020},
    pages       =   {5031-5041}
}

@inproceedings{lisa,
  author       = {Enric Corona and Tomas Hodan and Minh Vo and Francesc Moreno{-}Noguer and
                  Chris Sweeney and Richard A. Newcombe and Lingni Ma},
  title        = {{LISA:} Learning Implicit Shape and Appearance of Hands},
  booktitle    = CVPR,
  pages        = {20501--20511},
  year         = {2022}
}

@inproceedings{hand_meshes,
    author      =   {Luca Ballan and
                  Aparna Taneja and
                  J{\"{u}}rgen Gall and
                  Luc Van Gool and
                  Marc Pollefeys},
    title       =   {Motion Capture of Hands in Action Using Discriminative Salient Points},
    booktitle   =   ECCV,
    year        =   {2012},
    pages       =   {640-653},
}

@inproceedings{freihand,
    author       = {Christian Zimmermann and Duygu Ceylan and Jimei Yang and Bryan C. Russell and Max J. Argus and Thomas Brox},
    title       =   {{FreiHAND: A Dataset for Markerless Capture of Hand Pose and Shape From Single RGB Images}},
    booktitle   =   ICCV,
    year        =   {2019},
    pages       =   {813-822}
}

@inproceedings{fhpa,
    author      =   {Guillermo Garcia{-}Hernando and
                  Shanxin Yuan and
                  Seungryul Baek and
                  Tae{-}Kyun Kim},
    title       =   {First-Person Hand Action Benchmark With {RGB-D} Videos and {3D} Hand Pose Annotations},
    booktitle   =   CVPR,
    year        =   {2018},
    pages       =   {409-419}
}

@inproceedings{honnotate,
    author      =   {Shreyas Hampali and
                  Mahdi Rad and
                  Markus Oberweger and
                  Vincent Lepetit},
    title       =   {{HOnnotate}: {A} Method for {3D} Annotation of Hand and Object Poses},
    booktitle   =   CVPR,
    year        =   {2020},
    pages       =   {3196-3206}
}

@inproceedings{contactpose,
    author      =   {Samarth Brahmbhatt and
                  Chengcheng Tang and
                  Christopher D. Twigg and
                  Charles C. Kemp and
                  James Hays},
    title       =   {{ContactPose}: {A} Dataset of Grasps with Object Contact and Hand Pose},
    booktitle   =   ECCV,
    year        =   {2020},
    pages       =   {361-378},
}

@inproceedings{dexYCB,
    author      =   {Yu{-}Wei Chao and
                  Wei Yang and
                  Yu Xiang and
                  Pavlo Molchanov and
                  Ankur Handa and
                  Jonathan Tremblay and
                  Yashraj S. Narang and
                  Karl Van Wyk and
                  Umar Iqbal and
                  Stan Birchfield and
                  Jan Kautz and
                  Dieter Fox},
    title       =   {{DexYCB}: {A}Benchmark for Capturing Hand Grasping of Objects},
    booktitle   =   CVPR,
    year        =   {2021},
    pages       =   {9044-9053}
}

@inproceedings{h2o,
    author       = {Taein Kwon and Bugra Tekin and Jan St{\"{u}}hmer and Federica Bogo and Marc Pollefeys},
    title       =   {{H2O}: {T}wo Hands Manipulating Objects for First Person Interaction Recognition},
    booktitle   =   ICCV,
    year        =   {2021},
    pages       =   {10138-10148}
}

@inproceedings{h2o-3d,
    author      =   {Shreyas Hampali and
                  Sayan Deb Sarkar and
                  Mahdi Rad and
                  Vincent Lepetit},
    title       =   {Keypoint Transformer: {S}olving Joint Identification in Challenging Hands and Object Interactions for Accurate {3D} Pose Estimation},
    booktitle   =   CVPR,
    year        =   {2022},
    pages       =   {11090-11100}
}

@inproceedings{hoi4d,
    author       = {Yunze Liu and Yun Liu and Che Jiang and Kangbo Lyu and Weikang Wan and Hao Shen and Boqiang Liang and Zhoujie Fu and He Wang and Li Yi},
    title       =   {{HOI4D}: {A} {4D} Egocentric Dataset for Category-Level Human-Object Interaction},
    booktitle   =   CVPR,
    year        =   {2022},
    pages       =   {21013-21022}
}

@inproceedings{arctic,
    author      =   {Zicong Fan and
                  Omid Taheri and
                  Dimitrios Tzionas and
                  Muhammed Kocabas and
                  Manuel Kaufmann and
                  Michael J. Black and
                  Otmar Hilliges},
    title       =   {{ARCTIC}: {A} Dataset for Dexterous Bimanual Hand-Object Manipulation},
    booktitle   =   CVPR,
    year        =   {2023},
    pages       =   {12943-12954}
}

@inproceedings{bps,
    author       = {Sergey Prokudin and Christoph Lassner and Javier Romero},
    title       =   {{Efficient Learning on Point Clouds With Basis Point Sets}},
    booktitle   =   ICCV,
    year        =   {2019},
    pages       =   {4332-4341}
}

@inproceedings{smplx,
    author       = {Georgios Pavlakos and Vasileios Choutas and Nima Ghorbani and Timo Bolkart and Ahmed A. A. Osman and Dimitrios Tzionas and Michael J. Black},
    title       =   {Expressive Body Capture: {3D} Hands, Face, and Body From a Single Image},
    booktitle   =   CVPR,
    pages       =   {10975--10985},
    year        =   {2019}
}

@inproceedings{circle,
    author      =   {Jo{\~{a}}o Pedro Ara{\'{u}}jo and
                  Jiaman Li and
                  Karthik Vetrivel and
                  Rishi Agarwal and
                  Jiajun Wu and
                  Deepak Gopinath and
                  Alexander Clegg and
                  C. Karen Liu},
    title       =   {{CIRCLE}: {C}apture in Rich Contextual Environments},
    booktitle   =   CVPR,
    year        =   {2023},
    pages       =   {21211-21221}
}

@inproceedings{saga,
    title       =   {{SAGA}: {S}tochastic Whole-Body Grasping with Contact},
    author       = {Yan Wu and Jiahao Wang and Yan Zhang and Siwei Zhang and Otmar Hilliges and Fisher Yu and Siyu Tang},
    booktitle   =   ECCV,
    pages       =   {257--274},
    volume      =   {13666},
    year        =   {2022}
}

@inproceedings{hand_primitives,
  title         =   {Efficient model-based {3D} tracking of hand articulations using {K}inect.},
  author       = {Iason Oikonomidis and Nikolaos Kyriazis and Antonis A. Argyros},

  booktitle     =   BMVC,
  year          =   {2011},
  pages         =   {1-11}
}

@inproceedings{imos,
    title       =   {{IMoS}: Intent-Driven Full-Body Motion Synthesis for Human-Object Interactions},
    author      =   {Anindita Ghosh and
                  Rishabh Dabral and
                  Vladislav Golyanik and
                  Christian Theobalt and
                  Philipp Slusallek},
    booktitle   =   CGF,
    year        =   {2023}
}

@inproceedings{toch,
    title       =   {{TOCH}: {S}patio-Temporal Object-to-Hand Correspondence for Motion Refinement},
    author       = {Keyang Zhou and Bharat Lal Bhatnagar and Jan Eric Lenssen and Gerard Pons{-}Moll},
    booktitle   =   ECCV,
    volume      =   {13663},
    pages       =   {1-19},
    year        =   {2022}
}

@conference{taheri_3dv2024_grip,
    title     = {{GRIP}: Generating Interaction Poses Using Spatial Cues and Latent Consistency},
    author    = {Taheri, Omid and Zhou, Yi and Tzionas, Dimitrios and Zhou, Yang and Ceylan, Duygu and Pirk, Soren and Black, Michael J.},
    booktitle =  I3DV,
    pages     = {933--943},
    year      = {2024}
}

@inproceedings{contactgen,
  title={{ContactGen}: {G}enerative Contact Modeling for Grasp Generation},
  author={Liu, Shaowei and Zhou, Yang and Yang, Jimei and Gupta, Saurabh and Wang, Shenlong},
  booktitle=ICCV,
  pages = {20552--20563},
  year={2023}
}

@inproceedings{dexgraspnet,
  title={{DexGraspNet}: {A} Large-Scale Robotic Dexterous Grasp Dataset for General Objects Based on Simulation},
  author={Ruicheng Wang and Jialiang Zhang and Jiayi Chen and Yinzhen Xu and Puhao Li and Tengyu Liu and He Wang},
  booktitle=ICRA,
  year={2022},
  pages={11359-11366}
}

@inproceedings{hoidiffusion,
  title={{HOIDiffusion}: {G}enerating Realistic {3D} Hand-Object Interaction Data},
  author={Mengqi Zhang and Yang Fu and Zheng Ding and Sifei Liu and Zhuowen Tu and Xiaolong Wang},
  booktitle=CVPR,
  pages={8521-8531},
  year={2024}
}

@inproceedings{geneoh,
   title={{GeneOH} Diffusion: {T}owards Generalizable Hand-Object Interaction Denoising via Denoising Diffusion},
   author={Liu, Xueyi and Yi, Li},
   booktitle=ICLR,
   year={2024}
}

@inproceedings{graspxl,
  title={{GraspXL}: {G}enerating Grasping Motions for Diverse Objects at Scale},
  author={Hui Zhang and Sammy Christen and Zicong Fan and Otmar Hilliges and Jie Song},
  booktitle=ECCV,
  pages={386--403},
  volume={15084},
  year={2024}
}

@inproceedings{omomo,
  title={Object Motion Guided Human Motion Synthesis},
  author={Jiaman Li and Jiajun Wu and C. Karen Liu},
  booktitle=TOG,
  volume       = {42},
  number       = {6},
  pages        = {197:1--197:11},
  year={2023}
}

@inproceedings{hasson2019obman,
    author      = {Yana Hasson and G{\"{u}}l Varol and Dimitrios Tzionas and Igor Kalevatykh and Michael J. Black and Ivan Laptev and Cordelia Schmid},
    title       = {Learning Joint Reconstruction of Hands and Manipulated Objects},
    booktitle   = CVPR,
    pages       = {11807--11816},
    year        = {2019}
}

@inproceedings{diomataris2024wandr,
    title = {{WANDR}: {I}ntention-guided Human Motion Generation},
    author = {Diomataris, Markos and Athanasiou, Nikos and Taheri, Omid and Wang, Xi and Hilliges, Otmar and Black, Michael J.},
    booktitle = CVPR,
    pages = {927-936},
    year = {2024},
}

@article{nimble,
    author = {Li, Yuwei and Zhang, Longwen and Qiu, Zesong and Jiang, Yingwenqi and Li, Nianyi and Ma, Yuexin and Zhang, Yuyao and Xu, Lan and Yu, Jingyi},
    title = {{NIMBLE}: {A} Non-Rigid Hand Model with Bones and Muscles},
    volume = {41},
    number = {4},
    journal = TOG,
    articleno = {120},
    pages = {120:1--120:16},
    numpages = {16},
    year = {2022}
}

@inproceedings{swamy2023showme,
    author       = {Anilkumar Swamy and Vincent Leroy and Philippe Weinzaepfel and Fabien Baradel and Salma Galaaoui and Romain Br{\'{e}}gier and Matthieu Armando and Jean{-}S{\'{e}}bastien Franco and Gr{\'{e}}gory Rogez},
    title        = {{SHOWMe}: {B}enchmarking Object-agnostic Hand-Object {3D} Reconstruction},
    booktitle    = ICCV,
    pages        = {1927--1936},
    year         = {2023}
}

@inproceedings{fan2024hold,
    author       = {Zicong Fan and Maria Parelli and Maria Eleni Kadoglou and Xu Chen and Muhammed Kocabas and Michael J. Black and Otmar Hilliges},
    title        = {{HOLD}: {C}ategory-Agnostic {3D} Reconstruction of Interacting Hands and
                  Objects from Video},
    booktitle    = CVPR,
    pages        = {494--504},
    year         = {2024}
}

@inproceedings{jiang2021handobjectcc,
  title={Hand-Object Contact Consistency Reasoning for Human Grasps Generation},
  author={Hanwen Jiang and Shaowei Liu and Jiashun Wang and Xiaolong Wang},
  booktitle=ICCV,
  year={2021},
  pages={11087-11096}
}

@inproceedings{turpin2022graspddc,
  title={{Grasp'D}: {D}ifferentiable Contact-rich Grasp Synthesis for Multi-fingered Hands},
  author={Dylan Turpin and Liquang Wang and Eric Heiden and Yun-Chun Chen and Miles Macklin and Stavros Tsogkas and Sven J. Dickinson and Animesh Garg},
  booktitle=ECCV,
  volume       = {13666},
  pages        = {201--221},
  year={2022}
}

@inproceedings{marchingcubes,
author = {Lorensen, William E. and Cline, Harvey E.},
title = {Marching cubes: A high resolution 3D surface construction algorithm},
year = {1987},
booktitle = SIGG,
pages = {163–169},
}

@inproceedings{paschalidis2025cwgrasp,
  title     = {{3D} {W}hole-Body Grasp Synthesis with Directional Controllability},
  author    = {Paschalidis, Georgios and Wilschut, Romana and Anti\'{c}, Dimitrije and Taheri, Omid and Tzionas, Dimitrios},
  booktitle = I3DV,
  year      = {2025}
 }

@inproceedings{zhang2025bimart,
title = {BimArt: A Unified Approach for the Synthesis of 3D Bimanual Interaction with Articulated Objects},
author = {Zhang, Wanyue and Dabral, Rishabh and Golyanik, Vladislav and Choutas, Vasileios and
	Alvarado, Eduardo and Beeler, Thabo and Habermann, Marc and Theobalt, Christian},
year = {2025},
booktitle = CVPR,
}

@inProceedings{Liu_2022_CVPR,
 author    = {Liu, Yunze and Liu, Yun and Jiang, Che and Lyu, Kangbo and Wan, Weikang and Shen, Hao and Liang, Boqiang and Fu, Zhoujie and Wang, He and Yi, Li},
 title     = {HOI4D: A 4D Egocentric Dataset for Category-Level Human-Object Interaction},
 booktitle = CVPR,
 year      = {2022},
}

@inproceedings{liu2024taco,
  title={TACO: Benchmarking Generalizable Bimanual Tool-ACtion-Object Understanding},
  author={Liu, Yun and Yang, Haolin and Si, Xu and Liu, Ling and Li, Zipeng and Zhang, Yuxiang and Liu, Yebin and Yi, Li},
  booktitle = CVPR,
  year={2024}
}

@inproceedings{hao2024hand,
  title={Hand-centric motion refinement for 3d hand-object interaction via hierarchical spatial-temporal modeling},
  author={Hao, Yuze and Zhang, Jianrong and Zhuo, Tao and Wen, Fuan and Fan, Hehe},
  booktitle=AAAI,
  year={2024}
}

@inproceedings{ye2023ghop,
    author = {Ye, Yufei and Gupta, Abhinav and Kitani, Kris and Tulsiani, Shubham},
    title = {G-HOP: Generative Hand-Object Prior for Interaction Reconstruction and Grasp Synthesis},
    booktitle = CVPR,
    year = {2024}
}

@article{zhang2025manidext,
    title={Manidext: Hand-object manipulation synthesis via continuous correspondence embeddings and residual-guided diffusion},
    author={Zhang, Jiajun and Zhang, Yuxiang and An, Liang and Li, Mengcheng and Zhang, Hongwen and Hu, Zonghai and Liu, Yebin},
    booktitle=PAMI,
    year={2025},
}

@inproceedings{cha2024text2hoi,
  title={Text2HOI: Text-guided 3D Motion Generation for Hand-Object Interaction},
  author={Cha, Junuk and Kim, Jihyeon and Yoon, Jae Shin and Baek, Seungryul},
  booktitle=CVPR,
  year={2024}
}

@inproceedings{lee2024interhandgen,
title = {InterHandGen: Two-Hand Interaction Generation via Cascaded Reverse Diffusion},
author = {Lee, Jihyun and Saito, Shunsuke and Nam, Giljoo and Sung, Minhyuk and Kim, Tae-Kyun},
booktitle = CVPR,
year = {2024}
}

@inproceedings{MACS2024,
  author = {Shimada, Soshi and Mueller, Franziska and Bednarik, Jan and Doosti, Bardia 
            and Bickel, Bernd and Tang, Danhang and Golyanik, Vladislav 
            and Taylor, Jonathan and Theobalt, Christian and Beeler, Thabo},
  title = {MACS: Mass Conditioned 3D Hand and Object Motion Synthesis}, 
  booktitle = I3DV, 
  year = {2024}
}

@InProceedings{Zheng_2023_CVPR,
    author    = {Zheng, Juntian and Zheng, Qingyuan and Fang, Lixing and Liu, Yun and Yi, Li},
    title     = {CAMS: CAnonicalized Manipulation Spaces for Category-Level Functional Hand-Object Manipulation Synthesis},
    booktitle = CVPR,
    year      = {2023},
}

@inProceedings{zhang2024artigrasp,
  title={{ArtiGrasp}: Physically Plausible Synthesis of Bi-Manual Dexterous Grasping and Articulation},
  author={Zhang, Hui and Christen, Sammy and Fan, Zicong and Zheng, Luocheng and Hwangbo, Jemin and Song, Jie and Hilliges, Otmar},
  booktitle=I3DV,
  year={2024}
}

@article{huang2023dynamic,
  title={Dynamic handover: Throw and catch with bimanual hands},
  author={Huang, Binghao and Chen, Yuanpei and Wang, Tianyu and Qin, Yuzhe and Yang, Yaodong and Atanasov, Nikolay and Wang, Xiaolong},
  journal=CORL,
  year={2023}
}

@article{lee2024dextouch,
  title={Dextouch: Learning to seek and manipulate objects with tactile dexterity},
  author={Lee, Kang-Won and Qin, Yuzhe and Wang, Xiaolong and Lim, Soo-Chul},
  journal=RAL,
  year={2024},
}

@InProceedings{turpin2023fastgraspd,
author       = {Dylan Turpin and Tao Zhong and Shutong Zhang and Guanglei Zhu and Eric Heiden and Miles Macklin and Stavros Tsogkas and Sven Dickinson and Animesh Garg},
title        = {Fast-Grasp'D: Dexterous Multi-finger Grasp Generation Through Differentiable Simulation},
booktitle    = ICRA,
year         = {2023},
}

@article{wan2023unidexgrasp++,
  title={UniDexGrasp++: Improving Dexterous Grasping Policy Learning via Geometry-aware Curriculum and Iterative Generalist-Specialist Learning},
  author={Wan, Weikang and Geng, Haoran and Liu, Yun and Shan, Zikang and Yang, Yaodong and Yi, Li and Wang, He},
  journal=ICCV,
  year={2023} 
}

@inproceedings{wang2024cyberdemo,
title={CyberDemo: Augmenting Simulated Human Demonstration for Real-World Dexterous Manipulation},
author={Wang, Jun and Qin, Yuzhe and Kuang, Kaiming and Korkmaz, Yigit and Gurumoorthy, Akhilan and Su, Hao and Wang, Xiaolong},
booktitle=CVPR,
year={2024}
}

@article{wang2022dexgraspnet,
  title={DexGraspNet: A Large-Scale Robotic Dexterous Grasp Dataset for General Objects Based on Simulation},
  author={Wang, Ruicheng and Zhang, Jialiang and Chen, Jiayi and Xu, Yinzhen and Li, Puhao and Liu, Tengyu and Wang, He},
  journal=ICRA,
  year={2022}
}

@article{touch-dexterity,
title          = {Rotating without Seeing: Towards In-hand Dexterity through Touch },
author         = {Yin, Zhao-Heng and Huang, Binghao and Qin, Yuzhe and Chen, Qifeng and Wang, Xiaolong},
journal        = ICRA,
year           = {2023},
}

@inproceedings{murali2025graspgen,
  title     = {GraspGen: A Diffusion-based Framework for 6-DOF Grasping with On-Generator Training},
  author    = {Murali, Adithyavairavan and Sundaralingam, Balakumar and Chao, Yu-Wei and Yamada, Jun and Yuan, Wentao and Carlson, Mark and Ramos, Fabio and Birchfield, Stan and Fox, Dieter and Eppner, Clemens},
  booktitle = ICRA,
  year      = {2026},
}

@InProceedings{Muchen_LatentHOI,
    author    = {Li, Muchen and Christen, Sammy and Wan, Chengde and Cai, Yujun and Liao, Renjie and Sigal, Leonid and Ma, Shugao},
    title     = {LatentHOI: On the Generalizable Hand Object Motion Generation with Latent Hand Diffusion.},
    booktitle = CVPR,
    year      = {2025},
}

@article{ho2022classifier,
  title={Classifier-free diffusion guidance},
  author={Ho, Jonathan and Salimans, Tim},
  journal={arXiv preprint arXiv:2207.12598},
  year={2022}
}

@inproceedings{brooks2023instructpix2pix,
  title={Instructpix2pix: Learning to follow image editing instructions},
  author={Brooks, Tim and Holynski, Aleksander and Efros, Alexei A},
  booktitle=CVPR,
  year={2023}
}

@article{song2020denoising,
  title={Denoising diffusion implicit models},
  author={Song, Jiaming and Meng, Chenlin and Ermon, Stefano},
  journal=ICLR,
  year={2021}
}

@article{10.1145/3528223.3530091,
author = {She, Qijin and Hu, Ruizhen and Xu, Juzhan and Liu, Min and Xu, Kai and Huang, Hui},
title = {Learning high-DOF reaching-and-grasping via dynamic representation of gripper-object interaction},
year = {2022},
issue_date = {July 2022},
publisher = {Association for Computing Machinery},
address = {New York, NY, USA},
volume = {41},
number = {4},
issn = {0730-0301},
url = {https://doi.org/10.1145/3528223.3530091},
doi = {10.1145/3528223.3530091},
journal = {ACM Trans. Graph.},
month = jul,
articleno = {97},
numpages = {14}
}

@article{10.1145/2574860,
author = {Zhao, Xi and Wang, He and Komura, Taku},
title = {Indexing 3D Scenes Using the Interaction Bisector Surface},
year = {2014},
issue_date = {May 2014},
publisher = {Association for Computing Machinery},
address = {New York, NY, USA},
volume = {33},
number = {3},
issn = {0730-0301},
url = {https://doi.org/10.1145/2574860},
doi = {10.1145/2574860},
journal = {ACM Trans. Graph.},
month = jun,
articleno = {22},
numpages = {14}
}

@article{
cao2025joint,
title={Joint Diffusion for Universal Hand-Object Grasp Generation},
author={Jinkun Cao and Jingyuan Liu and Kris Kitani and Yi Zhou},
journal={Transactions on Machine Learning Research},
issn={2835-8856},
year={2025},
url={https://openreview.net/forum?id=TZ0ztsYR6x},
note={}
}

@INPROCEEDINGS{wei2024dro,
    author={Wei, Zhenyu and Xu, Zhixuan and Guo, Jingxiang and Hou, Yiwen and Gao, Chongkai and Cai, Zhehao and Luo, Jiayu and Shao, Lin},
    booktitle={2025 IEEE International Conference on Robotics and Automation (ICRA)}, 
    title={$\mathcal{D}(\mathcal{R}, \mathcal{O})$ Grasp: A Unified Representation of Robot and Object Interaction for Cross-Embodiment Dexterous Grasping}, 
    year={2025},
    volume={},
    number={},
    pages={4982-4988},
    doi={10.1109/ICRA55743.2025.11127754}
}



\end{document}